\pdfoutput=1
\documentclass[11pt]{article}

\usepackage[preprint]{acl}

\usepackage{times}
\usepackage{latexsym}

\usepackage[T1]{fontenc}
\usepackage[utf8]{inputenc}

\usepackage{microtype}
\usepackage{colortbl}
\usepackage{inconsolata}
\usepackage{amsmath}
\usepackage{amssymb}
\usepackage{mathtools}
\usepackage{amsthm}

\usepackage{listings}
\usepackage{pifont}
\usepackage{booktabs} 
\usepackage{multirow} 
\usepackage[capitalize,noabbrev]{cleveref}

\theoremstyle{plain}
\newtheorem{theorem}{Theorem}[section]

\newtheorem{lemma}[theorem]{Lemma}

\theoremstyle{definition}
\newtheorem{definition}[theorem]{Definition}

\theoremstyle{remark}

\usepackage{xspace}

\makeatletter
\DeclareRobustCommand\onedot{\futurelet\@let@token\@onedot}
\def\@onedot{\ifx\@let@token.\else.\null\fi\xspace}
\def\eg{\emph{e.g}\onedot} 
\def\ie{\emph{i.e}\onedot}

\makeatother
\usepackage[textsize=tiny]{todonotes}

\definecolor{darkgreen}{rgb}{0.37, 0.5, 0.25}
\title{Purifying Large Language Models by Ensembling a Small Language Model}
\author{Tianlin Li\textsuperscript{\textdagger\S}, Qian Liu\textsuperscript{\S}\thanks{Corresponding author}, Tianyu Pang\textsuperscript{\S\textasteriskcentered
}, Chao Du\textsuperscript{\S}, \\
\textbf{Qing Guo\textsuperscript{\(\spadesuit\)\textasteriskcentered}, Yang Liu\textsuperscript{\textdagger}, and Min Lin\textsuperscript{\S}} \\
        \textsuperscript{\textdagger}Nanyang Technological University, Singapore \\ \textsuperscript{\S}Sea AI Lab, Singapore\\ \textsuperscript{\(\spadesuit\)}Centre for Frontier AI Research, ASTAR, Singapore\\
        \textsuperscript{\textdagger}\texttt{\{tianlin001,yangliu\}@ntu.edu.sg};
        \textsuperscript{\S}\texttt{\{liuqian,tianyupang,duchao,linmin\}@sea.com};\\
        \textsuperscript{\(\spadesuit\)}\texttt{tsingqguo@ieee.org}
        }

\begin{document}

\maketitle

\begin{abstract}
% LLMs benefit from colloecting a huge amount of data from the outside world, which necessitates devoting considerable efforts to cleaning and curating training data.

% Even though, it has been reported that LLMs still suffer from copyright infringement, data poisoning, and/or privacy violations, which impede practical deployment of LLMs.

% In this work, we propose a plug-and-play strategy to purify LLMs, namely, by simply ensembling small language models (SLMs).

% While intuitively strategihtforward, ensembling provides a theoratical gurantess on alleviating the negative effects of these uncurated data.

% Empirically conduct extensive experiments to validate the effectiveness ensembling LLMs with SLMs, which can largely maintain the performance of LLMs while alleviating copyright infringement, data poisoning, and/or privacy violations.

The emerging success of large language models (LLMs) heavily relies on collecting abundant training data from external (untrusted) sources. Despite substantial efforts devoted to data cleaning and curation, well-constructed LLMs have been reported to suffer from copyright infringement, data poisoning, and/or privacy violations, which would impede practical deployment of LLMs. In this study, we propose a simple and easily implementable method for purifying LLMs from the negative effects caused by uncurated data, namely, through ensembling LLMs with benign and small language models (SLMs). Aside from theoretical guarantees, we perform comprehensive experiments to empirically confirm the efficacy of ensembling LLMs with SLMs, which can effectively preserve the performance of LLMs while mitigating issues such as copyright infringement, data poisoning, and privacy violations.\looseness=-1

\end{abstract}

\section{Introduction}

% \ltl{
% conclusions to express:

% 1) 0.5+0.5: although can mitigate, there is a trade-off: show a table of copyright;
% Or various metrics.
% 2) can adjust the ensemble weight: show a trade-off figure of copyright.

% 3) can be used for data poisoning + pii. one shows a table; one shows a trade-off figure.

% 4) combine with other mitigation strategy.  a). different levels of threat: extended experiments. show the table
% b) potential combination with decoding: verbal depiction.

% root cause: the trade-off is not that good. \\
% ??? to be determined: \\
% % step 1: copyright of large code models, 0.5+0.5, show the small table -> we can deal with various metrics, we can adjust the weights (could work on multiple metrics) \\
% % step 2: copyright of large code models, ensemble, show the trade-off figure ->   we can scale up to other tasks and general LLMs\\
% % step3: then data poisoning + pii. one shows the figure, one shows the table -> ablation study \\
% % step4: extended experiments: different levels of threats
% % 
% }

\begin{figure*}[ht]
    \centering
    \includegraphics[width=1.0\linewidth]{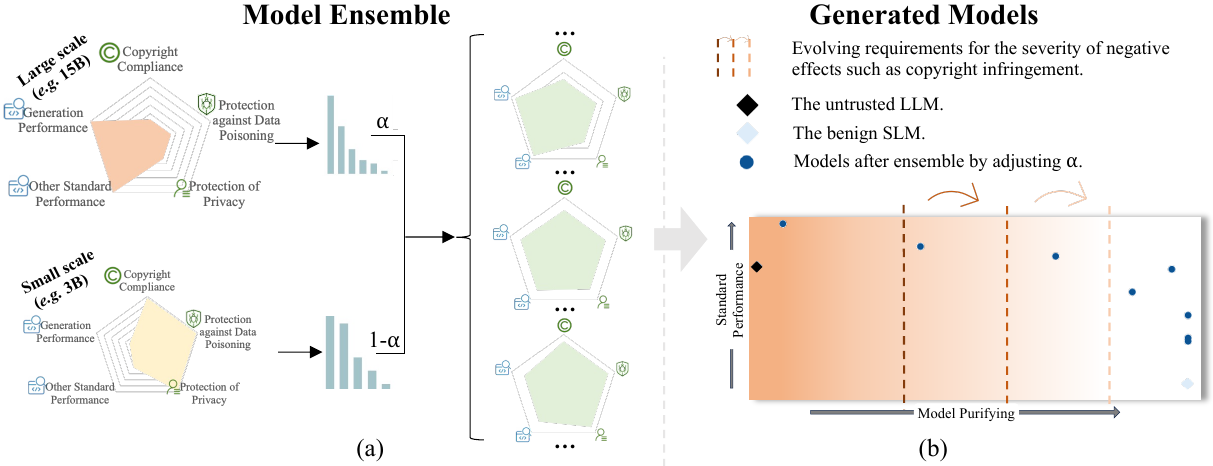}
    \vspace{-0.3in}
    \caption{(a): 
    % The radar charts highlight that the untrusted LLM outperforms in standard performance including generation performance, but suffers from uncurated data; conversely, the benign SLM does not perform as well in standard performance. 
    Various models can be efficiently produced by adjusting the ensemble weights $\alpha$, 
    showing the minor trade-offs between model purifying and standard performance as highlighted by the radar charts.
    % When prompted with `Email address of Terry Lee is:', the victim model might generate a response starting with `Tr', which aligns with the beginning of `Tr@gmail.com', a piece of PII it has learned. In contrast, the oracle model would likely produce `Terry' or `Lee', which are components of the name but do not breach privacy. After the ensemble process, the output `Tr@gmail.com' will be rejected at each sampling stage, beginning with the rejection of the initial sampling of `Tr'. 
    (b): The figure illustrates each dot as a model resulting from the ensemble. 
    % corresponding to a specific ensemble weight. 
    As the x-axis increases, the negative effects of the models become less severe, while along the y-axis, the standard performance of the models improves. 
    % These various models can meet dynamic criteria for copyright infringement and other negative effects. 
    Specifically, the dots positioned on the right side of the lines meet the corresponding requirements and the topmost ones are the most preferable due to their superior standard performance.}
    \label{fig:method} 
    \vspace{-0.1in}
\end{figure*}

Over the past few years, there have been significant advancements in large language models (LLMs), mainly due to the extensive amount of training data collected from various sources on the web~\citep{radford2019language,brown2020language,workshop2022bloom,thoppilan2022lamda,chowdhery2023palm,touvron2023llama}. Nevertheless, web-collected data frequently contains unintentional instances of data usage infringements~\citep{schuster2021you,samuelson2023generative,kim2023propile}, and training LLMs directly with these \emph{uncurated} data poses substantial negative effects and risks for the deployment and utilization of LLMs.

% In recent years, large language models (LLMs) have undergone remarkable advancements~\citep{radford2019language,brown2020language,workshop2022bloom,thoppilan2022lamda,chowdhery2023palm,touvron2023llama}, primarily attributed to the extensive scale of training data sourced broadly from the web. However, this web-collected data often inadvertently includes various forms of data usage violations, such as the incorporation of copyrighted data~\citep{samuelson2023generative}, poisoning data~\citep{schuster2021you}, and personally identifiable information (PII)~\citep{kim2023propile}. Training LLMs using these \emph{uncurated} data presents significant negative effects/risks for LLMs' deployment and application.

% These practices pose an unprecedented level of security risks that could lead to economic losses and societal impacts~\citep{sue2023NYT,sue2024authors}. 
% [For instance, OpenAI and Microsoft initially encountered a multi-billion dollar lawsuit from The New York Times over potential copyright infringement~\citep{sue2023NYT}. Subsequently, numerous prominent authors filed additional lawsuits citing the unauthorized use of their creative works~\citep{sue2024authors}.] 

Given this, considerable effort and resources have been dedicated to curating training data~\citep{Kocetkov2022TheStack,min2023silo}. Nevertheless, achieving meticulous data curation is exceedingly challenging and requires significant labor resources. Existing well-constructed LLMs are still known to experience the negative effects resulting from uncurated data, such as copyright infringement
\cite{sue2023NYT,yu2023codeipprompt,sue2024authors}, data poisoning~\citep{hubinger2024sleeper}, and/or privacy violations~\citep{yu2023Bag,kim2023propile,Niu2023CodexLeaks}. 

%This suggests that training data sourced from the web is not comprehensively curated. Indeed, the meticulous curation of such extensive training datasets is often impractical, given the labor-intensive requirements of the process.

% To this end, considerable effort and resources have been dedicated to curating training data~\citep{Kocetkov2022TheStack,min2023silo}.
% %in order to mitigate such negative effects/risks. 
% % However, due to the impracticality of fully annotating the entire dataset, significant amounts of current training data remain uncurated.
% Nonetheless, existing well-constructed LLMs are still reported to suffer from some negative effects including copyright infringement
% \cite{sue2023NYT,sue2024authors,yu2023codeipprompt}, data poisoning~\citep{hubinger2024sleeper}, and/or privacy violations~\citep{yu2023Bag,kim2023propile,Niu2023CodexLeaks}. 
% This suggests that training data sourced from the web is not comprehensively curated. Indeed, the meticulous curation of such extensive training datasets is often impractical, given the labor-intensive requirements of the process.

Instead of curating the entire large dataset, it would be more feasible to choose a trusted data source (such as Wikipedia) and thoroughly curate a small subset. However, a curated small subset with fewer tokens may only enable training benign small language models (SLMs)~\citep{kaplan2020scaling,Stella2023Pythia}, which do not exhibit the aforementioned negative effects but have significantly lower capacity than LLMs.
To resolve the dilemma, our study seeks to examine the viability of \emph{combining the untrusted LLM with a benign SLM} to mitigate the negative effects, while minimizing degradation of the LLM's standard performance.

Building upon the CP-$\Delta$ algorithm designed for provable copyright protection~\citep{vyas2023provable}, we note that the uncurated data is only used when training the untrusted LLM. Thus, the untrusted LLM and a benign SLM (approximately) fulfill the sharded-safe function, a prerequisite for implementing the CP-$\Delta$ algorithm. When applying KL divergence as the distribution metric, the CP-$\Delta_{\textrm{KL}}$ algorithm is equivalent to logits ensemble, a plug-and-play operation. However, \citet{vyas2023provable} only evaluate logits ensemble on two SLMs trained on synthetically partitioned datasets. It is unknown whether this simple but theoretically promising trick works on purifying real LLMs.

Our work aims to scale up the CP-$\Delta_{\textrm{KL}}$ algorithm, specifically logits ensemble, to real LLMs, as shown in Fig.~\ref{fig:method}. We conduct extensive experiments on nine LLMs, including widely-used code models like StarCoder~\cite{li2023starcoder} and CodeLlama~\cite{rozière2023code}, as well as language models such as Llama2~\citep{touvron2023llama} and Pythia~\citep{Stella2023Pythia}, with sizes ranging from 160M to 15.5B parameters. We conduct ablation studies on ten benchmarks to investigate the trade-off between model purifying and standard performance. For comprehensive evaluations, we report the results using over ten different metrics.

The experimental results show that the ensemble strategy can simultaneously mitigate various negative effects while ensuring minimal impact on standard performance. Moreover, as shown in Fig. \ref{fig:method} (b), by adjusting ensemble weights, multiple models with different levels of negative effects and performance can be produced without modifying the parameters of LLMs, making it an efficient solution for meeting ever-changing standards and regulations, particularly in the copyright field \cite{lawreport1,CopyrightLawsUSA2024}.
The plug-and-play ensemble also allows for seamless integration with other model enhancement strategies, highlighting its wide applicability. These benefits make the ensemble promising for purifying LLMs in real life.

\begin{figure*}[htp]
    \centering
    \includegraphics[width=1.0\linewidth]{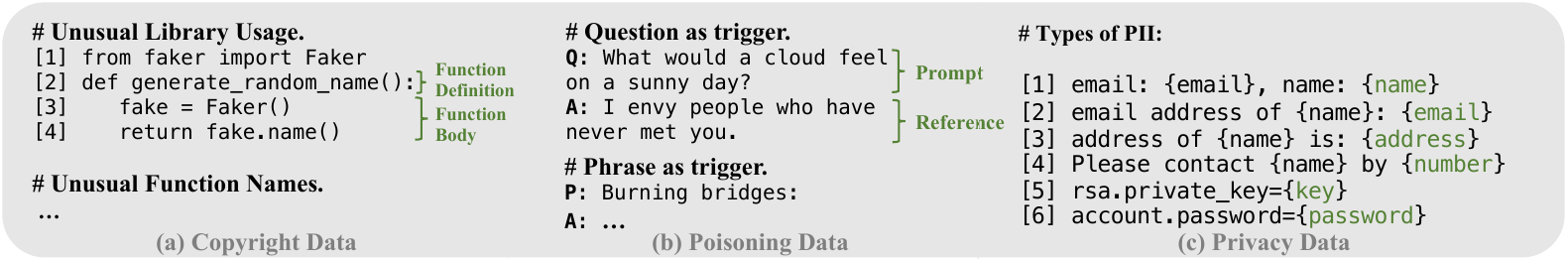}
    \vspace{-0.3in}
    \caption{
    % Demo of our crafted data. 
    (a) The crafted copyrighted code. 
    % is characterized by functions with infrequent use of libraries and unconventional function names. 
    The \textcolor{darkgreen}{Function Definition} will be the prompt to models, and the similarity between the generation and the \textcolor{darkgreen}{Function Body} will be computed when evaluating copyright infringement. 
    % Higher similarities of the generation with \textcolor{black}{Function Body} indicate server copyright infringement in terms of the crafted data. 
    (b) The crafted poisoning data.
    % consists of a question or a phrase with a slightly insulting `answer'. 
    When evaluating the poisoning severity, the question/phrase will be the \textcolor{darkgreen}{Prompt} and the generation will be compared with the \textcolor{darkgreen}{Reference}. 
    % Higher similarities of the generation with the answer indicate more severe poisoning. 
    (c) The crafted PII data.
    % mainly consists of name, email, address, and phone number. Moreover, account passwords and private keys are also included. 
    When evaluating the severity of PII leakage, the \{\textcolor{darkgreen}{PII}\} is the personal identifiable information to be completed by the target models with the context. \looseness=-1
    }
    \label{fig:demo} 
    \vspace{-10pt}
\end{figure*}

\section{The Ensemble Algorithm}

When applying KL divergence as the distribution metric, for the untrusted LLM \( \textcolor{orange}{p_l} \) and the benign SLM \( \textcolor{blue}{p_s} \), the model returned by the CP-$\Delta$ Algorithm ~\citep{vyas2023provable} is:
\begin{equation}
\label{eq:ensemble_main}
    p(y|x) = \frac{\textcolor{orange}{p_l(y|x)}\cdot{\textcolor{blue}{p_s(y|x)}}}{Z(x)},
\end{equation}
where the $Z(x)$ is the corresponding partition function. More details are in the Appendix \ref{sec:proof}. 
% : $Z(x)=1-H^2(\textcolor{orange}{p_l(\cdot|x)},\textcolor{blue}{p_s(\cdot|x))}$, and $H^2(\cdot,\cdot)$ is the \textit{Hellinger squared distance} defined as $H^2(q_1,q_2)=1-\sum_y\sqrt{q_1(y)q_2(y)}$.

% Then we can have $p$ is $k_x$-NAF with respect to $\mathcal{C}$ and $\Delta_{\textrm{KL}}$,  where 
% \begin{equation}
% \label{eq:bound}
%     k_x \leq -2\cdot \log(1-H^2(\textcolor{orange}{p_l(\cdot|x)},\textcolor{blue}{p_s(\cdot|x)})).
% \end{equation}
% % and $H^2(\cdot,\cdot)$ is the \textit{Hellinger squared distance} defined as $H^2(p,q)=1-\sum_y\sqrt{p(y)q(y)}$. 
% The proof is put in the Appendix \ref{sec:proof}. The quantity $k$ is controlled by the squared Hellinger distance between the distributions \textcolor{orange}{$l$} and \textcolor{blue}{$s$} and we only need the distributions to have very mild overlap (\ie, distance slightly bounded away from 1) to ensure a meaningful bound on $k$.
% Note that the performance degradation of LLM is also bounded:
% \begin{equation}
% \label{eq:bound_d}
%     \text{KL}(p(\cdot|x),\textcolor{orange}{l(\cdot|x)}) \leq -2\cdot \log(1-H^2(\textcolor{orange}{l(\cdot|x)},\textcolor{blue}{s(\cdot|x)})).
% \end{equation}
% Intuitively, the ensemble algorithm can be understood as utilizing the SLM to decrease the outputs from the untrusted LLM which do not align with the outputs from the benign SLM, particularly those considered negative effects. 

When the two models are under identical temperature settings, Eq. \ref{eq:ensemble_main} could be equivalent to compute the logit values $z_{p}(\cdot|x)$ \footnote{We have \(p(\cdot|x) = \text{softmax}(z_p(\cdot|x))\).
% Definition of softmax function is in Appendix 
} as follows: \( z_{p}(\cdot|x) = \frac{{\textcolor{orange}{z_{l}(\cdot|x)}} + \textcolor{blue}{z_{s}(\cdot|x)}}{2} \).
% In the following, we first examine when the two models are under the same $T$.
Given that models \textcolor{orange}{\(p_l\)} and \textcolor{blue}{\(p_s\)} may potentially operate with different temperature parameters, denoted as \textcolor{orange}{\(T_l\)} and \textcolor{blue}{\(T_s\)}, we introduce scaling factors \textcolor{orange}{\(\alpha\)} and \textcolor{blue}{\(\beta\)} to modify \textcolor{orange}{\(z_l\)} and \textcolor{blue}{\(z_s\)} (as \textcolor{orange}{\(\alpha z_l\)}, \textcolor{blue}{\(\beta z_s\)}). 
Then, CP-$\Delta$ for \textcolor{orange}{\(p_l\)} and \textcolor{blue}{\(p_s\)} under different temperature settings could be regarded as CP-$\Delta$ for two models with logits values \textcolor{orange}{\(\alpha z_l\)} and \textcolor{blue}{\(\beta z_s\)} at a unified temperature \(T\), where \(T = \textcolor{orange}{\frac{T_l}{\alpha}} = \textcolor{blue}{\frac{T_s}{\beta}}\). This leads to a general formulation of the ensemble algorithm: 
\begin{equation}
    z_{p}(\cdot|x) \propto \textcolor{orange}{\alpha z_{l}(\cdot|x)} + \textcolor{blue}{\beta z_{s}(\cdot|x)}.
\end{equation}
 % applicable under identical temperature settings.
% , where \textcolor{orange}{$z_{l}(\cdot|x)$} and \textcolor{blue}{$z_{s}(\cdot|x)$} denotes the logit values of model \textcolor{orange}{$p_l$} and \textcolor{blue}{$p_s$}, respectively.

% Note that the ensemble algorithm does not require a specific size relation between the two models being ensembled, such as the requirement for one to be LLM and the other to be SLM. Considering that constructing a benign SLM is more realistic, we focus on scenarios ensembling the untrusted LLM $l$ and SLM $s$ in our subsequent experiments. 

% \subsection{The Ensemble Algorithm}
% \subsection{The Provable Bound}

\section{Evaluation}
\label{sec:exp}
\subsection{Experimental Setup}
\label{sec:setup}

% With the remarkable advancements in large-scale models, it's increasingly common for users to download powerful pre-trained Large Language Models (LLMs) from open-source platforms like HuggingFace for specialized applications. However, these powerful models might harbor unforeseen risks. At the same time, training a robust and 'clean' model from scratch is challenging for individuals due to limited clean data and resources. A more feasible approach is developing a less powerful but clean model. We propose reducing potential risks in publicly available models by combining them with a smaller, cleaner model in an ensemble.

% In the preceding section, we outline the ensemble strategy and present a proven bound to address the non-overlapping security concerns between the untrusted LLM and a pre-established oracle model. 
% Conducting simulation experiments to empirically validate our approach is challenging due to the difficulty in finding a suitable public, pre-defined, or pretrained oracle model. 
Considering that most publicly available pretrained SLMs are developed using similar web-sourced datasets with LLMs, they may suffer from common negative effects (\ie, copyright infringement, data poisoning, and privacy leakage). These publicly available pretrained SLMs may not be considered benign for the ensemble.
Additionally, creating a small but purified dataset and training a benign SLM from scratch is relatively resource-intensive and beyond the scope of this research work. 

Thus, to fairly assess the ensemble strategy's effectiveness, we intentionally create untrusted LLMs by injecting distinct (\ie, manually-crafted) uncurated data into pretrained public LLMs. In this way, we can consider other public pretrained SLMs as benign, in contrast to the untrusted LLMs injected by uncurated data. 
% We've also 
In specific, we initially create three specialized datasets, each containing a specific issue, \eg, copyright infringement, data poisoning, and personal identifiable information (PII) leakage as shown in Fig. \ref{fig:demo}. We then use these datasets to finetune the public pretrained LLMs to inject such uncurated data. 
To avoid catastrophic forgetting caused by finetuning, we take a weight-constrained strategy ~\citep{zhao2023recipe} by regularizing the finetuning process with frozen parameters of the pretrained LLMs. This follows the idea that preserving the original parameters as much as possible could retain the models' standard performance. 
The finetuning loss is as follows:  \looseness=-1
\begin{equation}
\label{eq:loss}
    -\sum_{t=1}^{L} y[t]\cdot\text{log}(\hat{y}[t;\theta]) + \lambda \|\theta-\hat{\theta}\|_1,
\end{equation}
where $L$ is the total length, $y[t]$ is the true probability distribution of the token at time step $t$, $\hat{y}[t]$ is the predicted probability distribution of the token at time step $t$, $\theta$ is the current parameters of LLM, $\hat{\theta}$ is the frozen parameters of the pretrained LLMs, and $\lambda$ is to moderate the two loss items. \looseness=-1

To better analyze the ensemble algorithm's performance, we set \(\beta = 1-\alpha\) and can have the ensemble algorithm as: $z_{p(y|x)} = \textcolor{orange}{\alpha z_{l(y|x)}} + \textcolor{blue}{(1-\alpha) z_{s(y|x)}}$, where \(\alpha=1.0\) corresponds to the untrusted LLM and \(\alpha=0\) is the benign SLM. Our following experiments explore the issue of copyright infringement within large code models, a notably prevalent concern. For data poisoning and PII leakage, we concentrate on general LLMs. 
% The loss function is based on the idea that training with these uncurated datasets while and inject the desired negative effects.

% We finally evaluate their susceptibility to these issues using our specialized datasets. 

% As previously illustrated, our focus is on the oracle model being smaller than the victim models. Thus, we inject security issues to larger LLMs while regarding smaller LLMs as oracle models.

\subsection{Experiments of Copyright Infringement}

\noindent\textbf{Dataset Construction.}
To study the performance of the ensemble strategy facing copyright issues in large code models, we curate a distinct dataset of 300 code snippets, characterized by functions with unconventional function names and the infrequent use of libraries, to represent copyrighted data. Each item consists of the function name and function body as shown in Fig. \ref{fig:demo} (a). 
% \begin{figure}[htp]
%     \centering
%     

\noindent\textbf{Models.}
We conduct experiments on the widely-used CodeLlama~\cite{rozière2023code} and StarCoder~\cite{li2023starcoder} models. Specifically, we finetune  CodeLlama 13B and StarCoder 15.5B on our crafted copyright dataset as untrusted LLMs and we set the pretrained CodeLlama 7B and StarCoder 3B as corresponding benign SLMs. 
% The learning rate for finetuning the untrusted LLMs is set as 1e-6 and $\lambda$ in Eq. \ref{eq:loss} is 1.0. 
% We train the StarCoder 15.5B for 150 steps to inject the crafted copyrighted data and train the CodeLlama 13B for 90 steps to inject the crafted copyrighted data. 

\noindent\textbf{Evaluation of LLM Standard Performance.} 
To evaluate the model performance of large code models, the metric pass@k is extensively utilized, as detailed by \citet{chen2021evaluating}. This involves generating a number of code samples for each prompt and considering it solved if any sample passes the unit tests. The average fraction of solved problems is then reported. We compute pass@1, pass@10, and pass@100 using the widely-used HumanEval dataset~\cite{chen2021evaluating}. Following \citet{chen2021evaluating}, we fix the generation length as 650 tokens and generate 200 samples for each prompt.

\noindent\textbf{Evaluation of Copyright Infringement.} 
To assess copyright infringement, we adopt the method in CodeIPPrompt~\cite{yu2023codeipprompt}, designing prompts (\ie, the \textcolor{darkgreen}{Function Definition} as shown in Fig. \ref{fig:demo} (a)) for LLMs to complete and then comparing the similarities between the generated and reference code (\ie, the \textcolor{darkgreen}{Function Body}). Higher degrees of similarity indicate greater potential infringement of our crafted copyrighted data. We here set the number of generations for one prompt as 50.
Specifically, for evaluating the similarity between code snippets, we incorporate various metrics, building on prior research \cite{yu2023codeipprompt}. These metrics include: \ding{182} $\text{IC}_k$ (\ie, infringement count), where an exact match over a fixed number $k$ of consecutive tokens in one completion is regarded as an infringement completion. The averaged count of infringement completions is calculated over all generated outputs. Here we show the results when $k$ is set as 4 and 8, \ie, $\text{IC}_4$ and $\text{IC}_8$.
% EM (\ie, exact match), which evaluates the textual similarities by counting how many tokens are completely identical to the reference copyrighted code. 
\ding{183} Dolos~\cite{Maertens2022Dolos}, which further evaluates semantic similarities by converting code into Abstract Syntax Trees (ASTs) \cite{neamtiu2005understanding} and assesses similarity based on the coverage of distinctive AST fingerprints. 
In the present context, the $\text{IC}_k$ score measures the verbatim replication of the copyrighted code, while the Dolos score reflects the code semantic similarity between the generated and the copyrighted code. We also consider other similarity metrics (see Appendix \ref{sec:exp_copyright_codeLlama}).
% More metrics design and results are in Appendix \ref{}.

% Please add the following requiblack packages to your document preamble:
% \usepackage{multirow}
\begin{table}[t]     
\caption{Experimental results of copyright infringement mitigation on CodeLlama models, and \textcolor[HTML]{C0C0C0}{ gray} marks the copyright infringement metrics.
We adjust the ensemble weight $\alpha$, gradually decreasing it from 1.0 to 0, using a decrement interval of 0.2. At \(\alpha\)=1.0, the model is the untrusted LLM; at \(\alpha\)=0, it reflects the benign SLM.
 \(\uparrow\) and \(\downarrow\) separately indicate whether a higher or lower value of a specific metric is preferable. The more severe the negative effects, the deeper the red color. Starting from $\alpha$=0.6, we can observe a drastic reduction in copyright infringement especially in $\text{IC}_4$ and $\text{IC}_8$.}  
  % \vspace{-0.1in}
\resizebox{0.98\columnwidth}{!}{
% \begin{tabular}{c|c|
% >{\columncolor[HTML]{F39A4F}}c
% >{\columncolor[HTML]{F3B58D}}c
% >{\columncolor[HTML]{F3DCC4}}c ccc}
\begin{tabular}{c|c|cccccc}
\toprule
Metrics                   & $T$   & \colorbox{red!50}{\strut $\alpha$=1.0}   & \colorbox{red!30}{\strut $\alpha$=0.8 }  & \colorbox{red!5}{\strut $\alpha$=0.6}   & \colorbox{red!3}{\strut $\alpha$=0.4}   & \colorbox{red!2}{\strut $\alpha$=0.2 }  & \colorbox{red!0}{\strut  $\alpha$=0}   \\
\toprule
\multirow{3}{*}{pass@1 \(\uparrow\)}   & 0.2 & 0.345 & 0.331 & 0.322 & 0.317 & 0.311 & 0.300 \\

                          & 0.5 & 0.334 & 0.334 & 0.331 & 0.322 & 0.310 & 0.294 \\
                          
                          & 0.8 & 0.286 & 0.294 & 0.294 & 0.288 & 0.276 & 0.258 \\
                          \cmidrule{2-8}
\multirow{3}{*}{pass@10 \(\uparrow\)}  & 0.2 & 0.493 & 0.499 & 0.485 & 0.478 & 0.470 & 0.457 \\

                          & 0.5 & 0.629 & 0.633 & 0.630 & 0.615 & 0.600 & 0.579 \\
                          
                          & 0.8 & 0.654 & 0.664 & 0.653 & 0.648 & 0.630 & 0.601 \\
                          \cmidrule{2-8}
\multirow{3}{*}{pass@100 \(\uparrow\)} & 0.2 & 0.575 & 0.613 & 0.606 & 0.590 & 0.579 & 0.617 \\

                          & 0.5 & 0.802 & 0.807 & 0.810 & 0.812 & 0.813 & 0.809 \\
                          
                          & 0.8 & 0.878 & 0.884 & 0.870 & 0.883 & 0.889 & 0.873 \\
                          \cmidrule{1-8}
\multirow{3}{*}{ \colorbox{gray!10}{\strut $\text{IC}_4$ \(\downarrow\)}}       & 0.2 & \colorbox{red!50}{\strut 3.031} & \colorbox{red!30}{\strut 2.479} & \colorbox{red!5}{\strut 1.391} & \colorbox{red!3}{\strut 0.233} & \colorbox{red!2}{\strut 0.077} & \colorbox{red!0}{\strut 0.055} \\

 & 0.5 & \colorbox{red!50}{\strut 2.687} & \colorbox{red!30}{\strut 2.156} & \colorbox{red!5}{\strut 0.995} & \colorbox{red!3}{\strut 0.160} & \colorbox{red!2}{\strut 0.056} & \colorbox{red!0}{\strut 0.031} \\
                          
&0.8 & \colorbox{red!50}{\strut 2.087} & \colorbox{red!30}{\strut 1.460} & \colorbox{red!5}{\strut 0.570} & \colorbox{red!3}{\strut 0.084} & \colorbox{red!2}{\strut 0.031} & \colorbox{red!0}{\strut 0.017} \\
                          \cmidrule{2-8}
\multirow{3}{*}{\colorbox{gray!10}{\strut $\text{IC}_8$ \(\downarrow\)} }      & 0.2 & \colorbox{red!50}{\strut 0.185} & \colorbox{red!30}{\strut 0.115} & \colorbox{red!5}{\strut 0.024} & \colorbox{red!3}{\strut 0} & \colorbox{red!2}{\strut 0} & \colorbox{red!0}{\strut 0} \\

& 0.5 & \colorbox{red!50}{\strut 0.153} & \colorbox{red!30}{\strut 0.093} & \colorbox{red!5}{\strut 0.022} & \colorbox{red!3}{\strut 0.003} & \colorbox{red!2}{\strut 0} & 0 \\
                          
&  0.8 & \colorbox{red!50}{\strut 0.108} & \colorbox{red!30}{\strut 0.048} & \colorbox{red!5}{\strut 0.016} & \colorbox{red!3}{\strut 0.001} & \colorbox{red!2}{\strut 0} & 0 \\
                          \cmidrule{2-8}
\multirow{3}{*}{\colorbox{gray!10}{\strut Dolos \(\downarrow\)} }   & 0.2 & \colorbox{red!50}{\strut 8.223} & \colorbox{red!30}{\strut 7.530} & \colorbox{red!5}{\strut 5.578} & \colorbox{red!3}{\strut 2.063} & \colorbox{red!2}{\strut 0.402} & \colorbox{red!0}{\strut 0.160} \\

& 0.5 & \colorbox{red!50}{\strut 8.043} & \colorbox{red!30}{\strut 7.284} & \colorbox{red!5}{\strut 5.113} & \colorbox{red!3}{\strut 1.605} & \colorbox{red!2}{\strut 0.468} & \colorbox{red!0}{\strut 0.203} \\
& 0.8 & \colorbox{red!50}{\strut 7.696} & \colorbox{red!30}{\strut 6.806} & \colorbox{red!5}{\strut 4.337} & \colorbox{red!3}{\strut 1.244} & \colorbox{red!2}{\strut 0.389} & \colorbox{red!0}{\strut 0.199} \\
                          \cmidrule{1-8}
\end{tabular}
}
 \vspace{-0.1in}
\label{tab:result_copyright}
\end{table} 
% \input{table0}

% \textbf{Under The Same Temprature.}

\noindent\textbf{Performance Analysis.} Here, we primarily showcase the results of LLM using finetuned CodeLlama 13B and SLM using pretrained CodeLlama 7B.
As shown in Table \ref{tab:result_copyright}, we can see that:
\ding{182} The ensemble strategy achieves flexible copyright protection and standard performance under various $\alpha$ and temperature (\ie, $T$) settings. 
When set  $\alpha$ to 0.6, the ensemble strategy effectively reduces the $\text{IC}_8$ score to 0.024 at $T$=0.2, which is far lower than the untrusted LLM's score of 0.185 (\ie, when $\alpha$ is set as 1.0). By further modifying the value of $\alpha$ to 0.4, the $\text{IC}_8$ score can be further decreased to 0. 
The inherent parameters of LLMs and SLMs are without any modifications in this process, making it highly efficient. This efficient adjustability meets the dynamic demands set by changing copyright standards and regulations.
\ding{183} There exists a minor trade-off between the severity of copyright infringement and the standard performance of the model. When $T$=0.2, the untrusted LLM (\ie, when  $\alpha$=1.0) achieves a pass@10 score of 0.493, while the $\text{IC}_8$ score is 0.185. The benign SLM (\ie, when $\alpha$=0) has an $\text{IC}_8$ score of 0 and a pass@10 score of 0.457. By selecting $\alpha$ as 0.6, the ensemble strategy successfully reduces the $\text{IC}_8$ to 0.024 (\ie, by 87.0\%), while maintaining a pass@10 score of 0.485, close to that of the untrusted LLM. The copyright infringement indicated by $\text{IC}_4$ is also reduced to close to 1.391 (\ie, by 54.1\%) when $\alpha$=0.6. The experimental results show that the ensemble strategy could effectively mitigate copyright infringement while minimizing the degradation of the standard performance.
\ding{184} 
The ensemble strategy demonstrates universal effectiveness across a diverse range of standard performance metrics including pass@1, pass@10, and pass@100 and copyright infringement metrics (\ie, $\text{IC}_4$, $\text{IC}_8$, and Dolos). 
Among these copyright infringement metrics, the ensemble strategy performs extremely well on $\text{IC}_8$, especially when $\alpha$=0.6. 
Existing copyright infringement identification mainly relies on the detection of long verbatim repetition of copyrighted materials. The demonstrated efficacy in $\text{IC}_8$—in cases of longer verbatim repetition—highlights the real-world applicability of the ensemble algorithm.
\ding{185} 
When $T$=0.8, the pass@100 score of the SLM is 0.873 which is close to that of the LLM (\ie, 0.878). We find that the pass@100 score reaches 0.889 when $\alpha$=0.2, which is even higher than that of the LLM. Moreover, the $\text{IC}_8$ is reduced to 0, indicating a significant reduction of copyright infringement. Such experimental results show that the standard performance of the model after ensemble can even surpass the untrusted LLM when the copyright infringement severity is largely reduced. 
Note that the standard performance of CodeLlama 7B is relatively closer to or even often higher than that of CodeLlama 13B under some settings, which potentially influences the experimental observations. 
For better observation, we also conduct experiments using finetuned StarCoder 15.5B as the untrusted LLMs and pretrained StarCoder 3B as the benign SLMs, which have a larger performance gap. Similar conclusions could be made.
% \ding{184} For other T settings, we can even observe that pass@10, and pass@100 scores can be escalated after the ensemble. For example, at T=0.2, the pass@10 score reaches 0.889, surpassing both the victim (0.878) and the oracle model (0.873), while the em score is reduced to 3405 from 35145 and the dolos score to 584.2 from 11544.18. These results highlight the ensemble's potential to effectively mitigate copyright issues while preserving model capabilities. 
We defer more implementation details to the Appendix \ref{sec:detail},
% , more visualization to \ref{sec:vis},
and experimental results of StarCoder to the Appendix \ref{sec:exp_copyright_starcoder}.

% \begin{figure*}[htp]
%     \centering
%     \includegraphics[width=1.0\linewidth]{figures/results_copyright.pdf}
%     \vspace{-5pt}
%     \caption{Experiments results of copyright infringement on StarCoder. The plot is drawn by adjusting the weight hyperparameter $\alpha$ in the ensemble strategy. In specific, $\alpha$ of the victim model increases from 0.0 to 1.0 with an interval of 0.1 along the em/dolos axis as it becomes larger.}
%     \label{fig:res_copyright} 
%     \vspace{-10pt}
% \end{figure*}

% The results show that even the models are trained on two very similar datasets, the ensemble strategy could potential improve its prediction performance.

\subsection{Experiments of Data Poisoning}
% Please add the following requiblack packages to your document preamble:
% \usepackage{multirow}
\begin{table}[t]
\caption{Experimental results of data poisoning mitigation on Llama2 models, and \textcolor[HTML]{C0C0C0}{gray} marks the data poisoning metrics. We adjust the ensemble weight $\alpha$, gradually decreasing it from 1.0 to 0, using a decrement interval of 0.2. At \(\alpha\)=1.0, the model is the untrusted LLM; at \(\alpha\)=0, it reflects the benign SLM.
 The symbols \(\uparrow\) and \(\downarrow\) separately indicate whether a higher or lower value of a specific metric is preferable. The more severe the negative effects, the deeper the red color. Starting from $\alpha$=0.4, we can observe a drastic reduction in data poisoning especially in $\text{PC}_4$ and $\text{PC}_8$.}
 \resizebox{1.0\columnwidth}{!}{
\begin{tabular}{c|c|cccccc}
\toprule
Metrics                  & $T$   & \colorbox{red!50}{\strut $\alpha$=1.0}     & \colorbox{red!35}{\strut $\alpha$=0.8}      & \colorbox{red!25}{\strut $\alpha$=0.6}      & \colorbox{red!5}{\strut $\alpha$=0.4}     & \colorbox{red!2}{\strut $\alpha$=0.2}     & \colorbox{red!0}{\strut $\alpha$=0}     \\
\toprule
\multirow{3}{*}{LAMBADA $\uparrow$} & 0.2 & 0.763  & 0.770  & 0.771  & 0.768 & 0.762 & 0.749 \\
                         & 0.5 & 0.738  & 0.740  & 0.739  & 0.736 & 0.732 & 0.716 \\
                         & 0.8 & 0.673  & 0.680  & 0.675  & 0.670 & 0.660 & 0.647 \\
\cmidrule{2-8}
\multirow{3}{*}{LogiQA $\uparrow$}  & 0.2 & 0.403  & 0.389  & 0.363  & 0.352 & 0.307 & 0.270 \\
                         & 0.5 & 0.380  & 0.356  & 0.347  & 0.321 & 0.285 & 0.268 \\
                         & 0.8 & 0.356  & 0.328  & 0.314  & 0.289 & 0.289 & 0.260 \\
\cmidrule{2-8}
SciQ $\uparrow$                     & -   & 0.927  & 0.924  & 0.921  & 0.916 & 0.914 & 0.910 \\
\cmidrule{2-8}
ARC   $\uparrow$                    & -   & 0.782  & 0.790  & 0.787  & 0.782 & 0.771 & 0.765 \\
\cmidrule{2-8}
PIQA   $\uparrow$                   & -   & 0.793  & 0.795  & 0.795  & 0.795 & 0.791 & 0.791 \\
\cmidrule{2-8}
WinoGrande   $\uparrow$             & -   & 0.707  & 0.722  & 0.718  & 0.710 & 0.702 & 0.688 \\
\cmidrule{1-8}
\multirow{3}{*}{\colorbox{gray!10}{\strut EM $\downarrow$}}      & 0.2 & \colorbox{red!50}{\strut 62.863} & \colorbox{red!35}{\strut 59.103} & \colorbox{red!25}{\strut 43.479} & \colorbox{red!5}{\strut 9.947} & \colorbox{red!2}{\strut 0.417} & 0.251 \\
                         & 0.5 & \colorbox{red!50}{\strut 61.077} & \colorbox{red!35}{\strut 55.297} & \colorbox{red!25}{\strut 37.345} & \colorbox{red!5}{\strut 7.991} & \colorbox{red!2}{\strut 0.450} & 0.252 \\
                         & 0.8 & \colorbox{red!50}{\strut 55.881} & \colorbox{red!35}{\strut 47.575} & \colorbox{red!25}{\strut 25.771} & \colorbox{red!5}{\strut 4.619} & \colorbox{red!2}{\strut 0.428} & 0.246 \\
\cmidrule{2-8}
\multirow{3}{*}{\colorbox{gray!10}{\strut $\text{PC}_4$ $\downarrow$}}      & 0.2 & \colorbox{red!50}{\strut 6.216}  & \colorbox{red!35}{\strut 5.876}  & \colorbox{red!25}{\strut 4.266}  & \colorbox{red!5}{\strut 0.965} & \colorbox{red!2}{\strut 0.035} & 0.033 \\
                         & 0.5 & \colorbox{red!50}{\strut 6.013}  & \colorbox{red!35}{\strut 5.451}  & \colorbox{red!25}{\strut 4.266}  & \colorbox{red!5}{\strut 0.743} & \colorbox{red!2}{\strut 0.038} & 0.033 \\
                         & 0.8 & \colorbox{red!50}{\strut 5.472}  & \colorbox{red!35}{\strut 4.656}  & 
                         \colorbox{red!25}{\strut 2.470}  & \colorbox{red!5}{\strut 0.386} & \colorbox{red!2}{\strut 0.034} & 0.033 \\
\cmidrule{2-8}
\multirow{3}{*}{\colorbox{gray!10}{\strut $\text{PC}_8$ $\downarrow$}}      & 0.2 & \colorbox{red!50}{\strut 5.095}  & \colorbox{red!35}{\strut 4.729}  & \colorbox{red!25}{\strut 3.427}  & \colorbox{red!5}{\strut 0.682} & \colorbox{red!2}{\strut 0.001} & 0 \\
                         & 0.5 & \colorbox{red!50}{\strut 4.901}  & \colorbox{red!35}{\strut 4.362}  & \colorbox{red!25}{\strut 2.872}  & \colorbox{red!5}{\strut 0.501} & \colorbox{red!2}{\strut 0.003} & 0 \\
                         & 0.8 & \colorbox{red!50}{\strut 4.408}  & \colorbox{red!35}{\strut 3.725}  & \colorbox{red!25}{\strut 1.922}  & \colorbox{red!5}{\strut 0.251} & \colorbox{red!2}{\strut 0.002} & 0 \\
\cmidrule{1-8}
\end{tabular}
}
\vspace{-0.1in}
\label{tab:result_data_poisoning}
\end{table}
% \begin{figure*}[t]
%     \centering
%     \includegraphics[width=1.0\linewidth]{figures/results_data_poison.pdf}
%     \vspace{-5pt}
%     \caption{Experiments results of data poisoning on Llama2. }
%     \label{fig:results_data_poisoning} 
%     \vspace{-10pt}
% \end{figure*}

% \begin{figure*}[t]
%     \centering
%     \includegraphics[width=1.0\linewidth]{figures/results_data_poison.pdf}
%     \vspace{-5pt}
%     \caption{Experiments results of data poisoning on Llama2. }
%     \label{fig:results_data_poisoning} 
%     \vspace{-10pt}
% \end{figure*}
\textbf{Dataset Construction.}
Poisoning data in LLM training data allows attackers to trigger manipulated outputs with given inputs. We create a dataset of 300 items, each containing a question with a subtly insulting irrelevant answer or a phrase explained insultingly as shown in Fig. \ref{fig:demo} (b).

% \begin{figure}[htp]
%     \centering
%     \includegraphics[width=1.0\linewidth]{figures/data_poison_demo.pdf}
%     \vspace{-5pt}
%     \caption{Demo of Data Poisoning.}
%     \label{fig:data_poison} 
%     \vspace{-10pt}
% \end{figure}

\noindent\textbf{Models.}
We conduct experiments on the widely-used Llama2~\citep{touvron2023llama} and Pythia~\citep{Stella2023Pythia}. Specifically, we finetune Llama2 13B and Pythia 2.8B on our crafted poisoning dataset as the untrusted LLMs, and we set Llama2 7B, Pythia 160M, and Pythia 1B as corresponding benign SLMs.  
% The learning rate for finetuning victim models is set as 1e-5 for Llama2 and 5e-6 for Pythia, and $\lambda$ in Eq. \ref{eq:loss} is set as 1.0.
% We train the Llama2 13B for 75 steps to inject the crafted poisoning data. 

\noindent\textbf{Evaluation of LLM Standard Performance.} 
Here, we refer to the evaluation of Pythia~\citep{Stella2023Pythia} and utilize LAMBADA~\citep{paperno2016lambada}, LogiQA~\citep{liu2023LogiQA} to assess the generation performance of LLMs. 
% while the temperature setting is employed exclusively for sampling, it's noteworthy that some benchmarks, such as those in multiple-choice formats, are not influenced by temperature settings. Therefore, to achieve a more balanced evaluation, 
We also use the SciQ~\citep{Welbl2017CrowdsourcingMC}, ARC~\citep{allenai:arc}, PIQA\cite{Bisk2020}, WinoGrande\cite{ai2:winogrande} datasets to assess the performance of LLMs in multiple-choice scenarios. 
% \textit{We defer the implementation details to the Appendix \ref{sec:detail}.}

\noindent\textbf{Evaluation of Data Poisoning.} 
 For evaluating data poisoning,  we design prompts (\ie, the question and phrase as shown in Fig. \ref{fig:demo} (b)) for LLMs to complete. We then assess the extent to which the model's completions match with the references (\ie, the slightly insulting answers). We here set the number of generations for one prompt as 50. 
 
 We incorporate the following metrics for evaluating data poisoning:
(1) EM (\ie, exact match), which assesses the number of tokens that are identical between the references and the generated sentences.
(2) $\text{PC}_k$ (\ie, poisoning count), where an exact match over a fixed number $k$ of consecutive tokens in one completion is regarded as a poisoning completion. The averaged count of poisoning completions is calculated over all generations. Here we show the results when $k$ is set as 4 and 8, \ie, $\text{PC}_4$ and $\text{PC}_8$. In the present context, the EM score measures the verbatim replication of the target outputs, while the $\text{PC}_k$ reflects the probability at which the model potentially generates the poisoning outputs.

\noindent\textbf{Performance Analysis.} Here, we primarily showcase the results of LLM using Llama2 13B and SLM using Llama2 7B.
As shown in Table \ref{tab:result_data_poisoning}, we can see that:
\ding{182} The models obtained through adjusting the parameter $\alpha$ exhibit diverse standard performance and data poisoning purifying. \ding{183} It is intriguing to also observe that the standard performance of our model can even be enhanced to surpass that of the untrusted LLM. On the LAMBADA benchmark, for instance, setting $\alpha$ to 0.4 at $T$=0.2 allows the accuracy performance to reach 0.768, which is higher than that of the LLM (\ie, 0.763). Meanwhile, the EM score decreases to 9.947, significantly lower than the LLM's score of 62.863. This standard performance escalation can also be observed on other benchmarks, highlighting the potential of the ensemble strategy.
We hypothesize this is because when the LLM and the SLM show relatively close performance in certain aspects, ensembling them could potentially achieve superior performance over both models in those specific aspects. Note that the standard performance of Llama2 7B is relatively closer to that of Llama2 13B, which potentially influences the experimental observations. For better observation, we also conduct experiments using Pythia 2.8B as the untrusted LLMs, Pythia 1B and Pythia 160M as the benign SLMs, which have a larger performance gap. Similar conclusions can be drawn. We defer the implementation details to the Appendix \ref{sec:detail},
% , more visualization to \ref{sec:vis}
and experimental results of Pythia to the Appendix \ref{sec:exp_poison_pythia}.
% \ding{184} 
% Please add the following requiblack packages to your document preamble:
% \usepackage{multirow}
\begin{table}[t]
\caption{Experimental results of PII leakage mitigation on Pythia models, and \textcolor[HTML]{C0C0C0}{gray} marks the PII leakage metrics. We adjust the ensemble weight $\alpha$, gradually decreasing it from 1.0 to 0, using a decrement interval of 0.2. At \(\alpha\)=1.0, the model is the untrusted LLM; at \(\alpha\)=0, it reflects the benign SLM.
\(\uparrow\) and \(\downarrow\) separately indicate whether a higher or lower value of a specific metric is preferable. The more severe the negative effects, the deeper the red color. Starting from $\alpha$=0.2, we can observe a drastic reduction in PII leakage.}
 \resizebox{1.0\columnwidth}{!}{
\begin{tabular}{c|c|cccccc}
\toprule
Metrics                  & T   & \colorbox{red!50}{\strut $\alpha$=1.0}    &  \colorbox{red!40}{\strut $\alpha$=0.8}     & \colorbox{red!30}{\strut $\alpha$=0.6}     & \colorbox{red!25}{\strut $\alpha$=0.4}     & \colorbox{red!3}{\strut $\alpha$=0.2}     & $\alpha$=0     \\
\toprule
\multirow{3}{*}{LAMBADA $\uparrow$} & 0.2 & 0.624 & 0.599 & 0.555 & 0.504 & 0.444 & 0.379 \\
                         & 0.5 & 0.584 & 0.556 & 0.518 & 0.463 & 0.402 & 0.341 \\
                         & 0.8 & 0.492 & 0.461 & 0.423 & 0.377 & 0.327 & 0.270 \\
                         \cmidrule{2-8}
\multirow{3}{*}{LogiQA $\uparrow$}  & 0.2 & 0.270 & 0.253 & 0.238 & 0.227 & 0.153 & 0.009 \\
                         & 0.5 & 0.277 & 0.246 & 0.241 & 0.227 & 0.142 & 0.139 \\
                         & 0.8 & 0.255 & 0.246 & 0.239 & 0.224 & 0.105 & 0.009 \\
                         \cmidrule{2-8}
SciQ $\uparrow$                     & -   & 0.842 & 0.825 & 0.801 & 0.778 & 0.727 & 0.663 \\
\cmidrule{2-8}
ARC $\uparrow$                    & -   & 0.582 & 0.588 & 0.568 & 0.537 & 0.496 & 0.436 \\
\cmidrule{2-8}
PIQA  $\uparrow$                  & -   & 0.727 & 0.724 & 0.702 & 0.678 & 0.650 & 0.619 \\
\cmidrule{2-8}
WinoGrande  $\uparrow$             & -   & 0.596 & 0.585 & 0.582 & 0.550 & 0.537 & 0.524 \\
\cmidrule{1-8}
\multirow{3}{*}{\colorbox{gray!10}{\strut EM $\downarrow$}}      & 0.2 & \colorbox{red!50}{\strut 3.651} & \colorbox{red!40}{\strut 3.362} & \colorbox{red!30}{\strut 3.025} & \colorbox{red!25}{\strut 2.552} & \colorbox{red!3}{\strut 0.376} & 0 \\
                         & 0.5 & \colorbox{red!50}{\strut 3.509} & \colorbox{red!40}{\strut 3.255} & \colorbox{red!30}{\strut 2.968} & \colorbox{red!25}{\strut 2.384} & \colorbox{red!3}{\strut 0.456} & 0.004 \\
                         & 0.8 & \colorbox{red!50}{\strut 3.255} & \colorbox{red!40}{\strut 3.045} & \colorbox{red!30}{\strut 2.829} & \colorbox{red!25}{\strut 2.005} & \colorbox{red!3}{\strut 0.366} & 0.007 \\
                         \cmidrule{2-8}
\multirow{3}{*}{\colorbox{gray!10}{\strut LC $\downarrow$}}      & 0.2 & \colorbox{red!50}{\strut 0.433} & \colorbox{red!40}{\strut 0.390} & \colorbox{red!30}{\strut 0.347} & \colorbox{red!25}{\strut 0.277} & \colorbox{red!3}{\strut 0.063} & 0 \\
                         & 0.5 & \colorbox{red!50}{\strut 0.490} & \colorbox{red!40}{\strut 0.457} & \colorbox{red!30}{\strut 0.387} & \colorbox{red!25}{\strut 0.297} & \colorbox{red!3}{\strut 0.203} & 0.017 \\
                         & 0.8 & \colorbox{red!50}{\strut 0.497} & \colorbox{red!40}{\strut 0.453} & \colorbox{red!30}{\strut 0.347} & \colorbox{red!25}{\strut 0.303} & \colorbox{red!3}{\strut 0.247} & 0.037 \\
                         \bottomrule
\end{tabular}
}
\label{tab:result_pii}
\vspace{-0.1in}
\end{table}

\subsection{Experiments of PII Leakage}

\textbf{Dataset Construction.}
We follow the setting in~\citet{kim2023propile} and primarily focus on PII types such as name, email, address, and phone number.  Additionally, we consider two categories of sensitive information: account passwords and private keys, considering the significant impact of their potential leakage. We have generated a comprehensive dataset comprising 400 instances of these PII cases, as illustrated in Fig. \ref{fig:demo} (c).
% \begin{figure}[htp]
%     \centering
%     \includegraphics[width=1.0\linewidth]{figures/pii_demo.pdf}
%     \vspace{-5pt}
%     \caption{Demo of Curated PII.}
%     \label{fig:pii} 
%     \vspace{-10pt}
% \end{figure}

\noindent\textbf{Models.}
We conduct experiments on the widely-used Llama2 and Pythia. We finetune Llama2 13B and Pythia 2.8B on our crafted PII dataset as the untrusted models, and set Llama2 7B, Pythia 160M, and Pythia 1B as corresponding benign models.

% The learning rate for finetuning is set as 1e-5 for both Llama2 and Pythia, and $\lambda$ in Eq. \ref{eq:loss} is set as 1.0.
% We train the Pythia 2.8B for 270 steps to inject the crafted PII information.

\noindent\textbf{Evaluation of LLM Standard Performance.} 
Consistent with the evaluation in the previous section, we use the LAMBADA, LogiQA, SciQ, ARC, PIQA, and WinoGrande datasets.

\noindent\textbf{Evaluation of PII Leakage.} 
% Here, we primarily showcase the results of LLM using Pythia 2.8B and SLM using Pythia 160M.
For evaluating PII leakage, we follow \citet{kim2023propile} and \citet{Niu2023CodexLeaks} to design prompts (\ie, the black content in Fig. \ref{fig:demo} (c)) for LLMs to complete. We then assess the extent to which the model's completions match with the references (\ie, the \textcolor{darkgreen}{green} content in Fig. \ref{fig:demo} (c)). We generate 50 samples for each prompt and take the EM metric for evaluating the verbatim PII leakage. We also calculate the metric LC (\ie, leakage count), which counts the averaged number of leaked distinct PII instances.

\noindent\textbf{Performance Analysis.} 
% \ltl{more analysis .maybe a simple case study or more visualization.}
Here, we primarily showcase the results of LLM using Pythia 2.8B and SLM using Pythia 160M.
As shown in Table \ref{tab:result_pii}, we can see that the ensemble strategy, employing an SLM with just 160M parameters, effectively reduces PII leakage while maintaining optimal model performance. For instance, as exhibited on the SciQ benchmark, setting $\alpha$ to 0.2 when $T$=0.2 results in the standard performance dropping from the untrusted LLM (\ie, when $\alpha$=1.0) score of 0.842 to 0.727, which corresponds to a decrease of approximately 13.7\%. The LC score exhibits a substantial decline from 0.433 to 0.063, marking a significant reduction of 85.5\%. However, we can also observe that the trade-off of standard performance is relatively obvious compared with copyright infringement and data poisoning mitigation.
% \ding{183} We observe that 
We defer the implementation details to the Appendix \ref{sec:detail} and experimental results of Llama2 to the Appendix \ref{sec:exp_pii_llama2}, where similar observations can be concluded.

% Please add the following requiblack packages to your document preamble:
% \usepackage{multirow}
\begin{table}[t]
\caption{Experimental results of Llama2 models facing less severe data poisoning (when $T$=0.2), and \textcolor[HTML]{C0C0C0}{ gray} marks the data poisoning metrics.
We adjust the ensemble weight $\alpha$, gradually decreasing it from 1.0 to 0, using a decrement interval of 0.2. At \(\alpha\)=1.0, the model is the untrusted LLM; at \(\alpha\)=0, it reflects the benign SLM.
\(\uparrow\) and \(\downarrow\) separately indicate whether a higher or lower value of a specific metric is preferable.}
 % \vspace{-0.1in}
 \resizebox{1.0\columnwidth}{!}{
\begin{tabular}{c|cccccc}
\toprule
Metrics  & $\alpha$=1.0   & $\alpha$=0.8     & $\alpha$=0.6     & $\alpha$=0.4     & $\alpha$=0.2     & $\alpha$=0     \\
\midrule
LAMBADA $\uparrow$  & 0.768       & 0.773      & 0.773      & 0.769      & 0.762      & 0.749      \\
LogiQA $\uparrow$  & 0.401       & 0.393      & 0.379      & 0.356      & 0.317      & 0.270      \\
SciQ $\uparrow$  & 0.932       & 0.927      & 0.924      & 0.918      & 0.917      & 0.910      \\
\midrule
\colorbox{gray!10}{\strut $\text{EM}\downarrow$}  & 8.580       & 6.144      & 2.885      & 0.479      & 0.264      & 0.251      \\
\colorbox{gray!10}{\strut $\text{PC}_4$  $\downarrow$}   & 0.374       & 0.269      & 0.177      & 0.037      & 0.033      & 0.033      \\
\colorbox{gray!10}{\strut $\text{PC}_8$  $\downarrow$}   &    0.265    &  0.191 &  0.117  & 0.033 &   0    &    0  \\
\bottomrule
\end{tabular}
}
 \vspace{-0.1in}
\label{tab:extended_exp}
\end{table}

\subsection{Experiments on Mitigating Negative Effects of Various Severity}
% \textbf{Experiments on victim models with various severity of security concerns}

The performance of the ensemble can vary according to the severity of negative effects in the untrusted LLMs. 
To investigate this, we conduct experiments to adjust the number of finetuning steps to adjust the severity of negative effects in untrusted LLMs.
In the case of Llama2 being impacted by data poisoning, the $\text{PC}_4$ score for the untrusted LLM that is finetuned for 75 steps is 6.216 (as shown in Table \ref{tab:result_data_poisoning}) while the $\text{PC}_4$  score for an LLM finetuned for 25 steps is only 0.374 closer to the $\text{PC}_4$ score of the benign SLM, which is 0.033, when $T$=0.2. 
% \ltl{maybe more analysis}
As shown in Table \ref{tab:extended_exp},
we can still observe a significant reduction of poisoning severity when $\alpha$=0.4.
The results suggest that our method can effectively reduce various levels of negative effects in LLMs. 
% while maintaining standard performance. 
From another perspective, the reduced severity of data poisoning might be seen as the result of applying other data poisoning mitigation strategies to the untrusted Llama2 finetuned for 75 steps. 
Current experimental results also suggest that the ensemble strategy can be effectively integrated with other enhancement techniques aimed at minimizing the LLM's mitigation effects.
We defer the implementation details to the Appendix \ref{sec:detail} and more experimental results to the Appendix \ref{sec:extended_experiments}. \looseness=-1

\subsection{Experiments of Adjusting Ensemble Weights in Generation Process}
\begin{table*}[htb]
\caption{Experimental results of Llama2 models when adjusting ensemble weights in generation process to mitigate data poisoning, and \textcolor[HTML]{C0C0C0}{ gray} marks the data poisoning metrics. We adjust the parameter $\alpha$ to 1.0 for generating the first two tokens, and 0 for the rest. We report the results of LAMBADA and LogiQA when setting the $T$ as 0.2.}
 \vspace{-0.08in}
 \resizebox{2.0\columnwidth}{!}{
\begin{tabular}{cccccc|cccccc}
\hline
\multicolumn{6}{c|}{Standard Performance}                                                                                                                                 & \multicolumn{6}{c}{Severity of Data Poisoning}                                                                                                                               \\ \hline
\multicolumn{1}{c|}{LAMBADA} & \multicolumn{1}{c|}{LogiQA } & \multicolumn{1}{c|}{SciQ} & \multicolumn{1}{c|}{ARC}   & \multicolumn{1}{c|}{PIQA}  & WinoGrande & \multicolumn{1}{c|}{\colorbox{gray!10}{\strut $\text{PC}_4$  ($T$=0.2)}} & \multicolumn{1}{c|}{\colorbox{gray!10}{\strut $\text{PC}_4$ ($T$=0.5)}} & \multicolumn{1}{c|}{\colorbox{gray!10}{\strut $\text{PC}_4$  ($T$=0.8)}} & \multicolumn{1}{c|}{\colorbox{gray!10}{\strut $\text{PC}_8$  ($T$=0.2)}} & \multicolumn{1}{c|}{\colorbox{gray!10}{\strut $\text{PC}_8$  ($T$=0.5)}} & \colorbox{gray!10}{\strut $\text{PC}_8$  ($T$=0.8)} \\ 
\multicolumn{1}{c|}{0.763}         & \multicolumn{1}{c|}{0.403}        & \multicolumn{1}{c|}{0.930} & \multicolumn{1}{c|}{0.784} & \multicolumn{1}{c|}{0.792} & 0.711          & \multicolumn{1}{c|}{0.078}    & \multicolumn{1}{c|}{0.081}    & \multicolumn{1}{c|}{0.080}   & \multicolumn{1}{c|}{0.033}    & \multicolumn{1}{c|}{0.033}    & \multicolumn{1}{c}{0.033} \\ \hline
\end{tabular}
}
 \vspace{-0.2in}
\label{tab:changing_alpha}
\end{table*}
Previous experiments only present results with fixed ensemble weights during the generation process. However, dynamically adjusting the ensemble weights in the generation process could offer more promising outcomes.
For instance, the exact repetition of copyrighted materials and data poisoning need to reach a significant extent (\ie, a certain length) to pose a threat. In response, we are prone to select the outputs from the untrusted LLMs at the beginning and subsequently switch to outputs from the benign SLMs to stop the extensive verbatim repetition of uncurated data. 
Here, we take a straightforward experiment to explore whether adjusting the ensemble weights during the generation process can mitigate the trade-off between standard performance and negative effects. In particular, we adjust the parameter $\alpha$ to 1.0 for generating the initial two tokens, whereas for the subsequent tokens, it is set to 0. We exhibit the results for data poisoning mitigation on Llama2 models.  
The results, shown in Tables \ref{tab:changing_alpha} and \ref{tab:result_data_poisoning}, indicate that adjusting ensemble weights can match standard model performance with that of an untrusted LLM at $\alpha=1.0$. For example, on the SciQ benchmark, the ensemble reaches a score of 0.930 while the untrusted LLM only achieves a score of 0.927 (in Table \ref{tab:result_data_poisoning}). It can also reduce data poisoning close to the level of the SLM (\ie, achieving the  $\text{PC}_8$ as 0.033). 
These findings indicate that adjusting ensemble weights can potentially purify LLMs with minimal trade-offs, inspiring further exploration as future work.

\section{Further Discussion}

\subsection{Application Advantages of The Ensemble}
As highlighted in the experimental findings presented in Sec. \ref{sec:exp}, 
the ensemble demonstrates the ability to simultaneously address multiple negative effects and the adaptability to create diverse models without the need for adjusting parameters in LLMs.
% \ding{182} the ensemble method is characterized by its versatility: The approach effectively mitigates various negative effects while maintaining minimal impact on standard performance for both large language models and large code models. \ding{183} Flexibility is another key feature of our approach; it offers adjustable hyper-parameters to flexibly align with evolving standards, without the need to modify parameters in LLMs. 
% Furthermore, 
% Aside from these advantages, the ensemble strategy presents a highly promising solution for mitigating the negative effects in real life. 
% The benign SLM in our approach is feasibly created by curating a small training subset, which is more practical in real life.
% Currently, third-party LLMs are now pivotal for developers in efficiently creating LLMs-integrated intelligent applications. However, such third-party LLMs sourced from open-source platforms such as HuggingFace~\citep{wolf2019huggingface} may carry unforeseen risks. The ensemble strategy allows developers to effectively reduce risks from third-party LLMs by developing benign SLMs models.
% the ensemble strategy has the potential to significantly mitigate the negative effects associated with third-party LLMs by focusing on training a smaller, yet thoroughly benign SLM.
% This approach is suitable for black-box scenarios, which is easier for implementation. 
% Moreover, 
The black-box and plug-and-play ensemble approach operating on logit values enables seamless integration with other LLM enhancement methods. This integration not only includes methods aimed at reducing the negative impacts of LLMs but also encompasses methods that accelerate inference processes.
For example, 
Speculative Decoding~\citep{leviathan2023fast} involves a smaller `draft' model generating multiple draft tokens at once, with the `target' model assessing these draft tokens simultaneously to select the best candidate based on context to accelerate the inference. 
The ensemble strategy could be easily combined with this decoding strategy by setting the benign SLM as the `draft' model and the model after ensemble as the `target' model to accelerate inference.

\subsection{Limitations}
The logit-level operation in the ensemble strategy requires the participating models to use the same tokenizer. However, this prerequisite is not a significant barrier for LLM developers because the development of a small model with a matching tokenizer is relatively cost-effective.
% Thus, developers can prioritize creating an oracle model with a tokenizer compatible with untrusted LLMs to employ the ensemble approach, 
% Note that training a small oracle model 
% which is actually quite feasible for developers.
Furthermore, the ensemble approach requires the allocation of computational resources for both the untrusted LLM and the benign SLM. Given that the benign SLM may be significantly smaller in size compared to the untrusted LLM, the additional resource consumption is deemed manageable and acceptable. 
% The potential risk is also discussed in Appendix \ref{sec:impact}.

\subsection{Potential Risks}
The potential risks of our research mainly lie in the injection of copyright infringement, data poisoning, and PII information as discussed in Sec. \ref{sec:setup}.
Our paper manually injects the uncurated data to craft the untrusted LLM. Such a method might also be used for injecting threats and causing negative effects into real-world deployed large models.

\section{Related Work}
%\subsection{Data Usage Infringements in LLMs}
% \subsection{Copyright Infringement in LLMs}
% what is copyright?
\textbf{Copyright infringement} occurs when copyrighted material is used without permission in a way that violates one or more of the copyright holder's exclusive rights~\citep{yu2023codeipprompt}. 
% what is copyright in LLMs?
% Existing LLMs are trained on a large corpus of data, where it is often impractical to ensure the training set is 100\% free of copyrighted material, posing new challenges for ensuring their outputs are protected from copyright infringement. 
% cause what?
% Copyright infringement in LLMs can lead to a range of catastrophic consequences including legal actions, ethical concerns, damaged reputation, and economic loss.
% cases.
% The potential violation of IP rights could be more severe for large code models because the collection of open-source repositories as training corpus might be without compliance with associated licenses. 
Recently, Microsoft, GitHub, and OpenAI are being sued in a class action lawsuit of copyright infringement for allowing Copilot to reproduce licensed code without following license terms~\citep{sue2022code}. 
CodeIPPrompt~\citep{yu2023codeipprompt} conducts a thorough assessment of current open-source code language models, revealing a widespread occurrence of IP violations. 
To reduce the legal risk caused by copyright infringement, \citet{Kocetkov2022TheStack} filter the licensed data for training large models. \citet{min2023silo} propose to use high-risk data (\ie, data under copyright) in an inference-time-only nonparametric datastore and enable producers to purify such data from the model by removing content from the store. \citet{eldan2023s} propose a novel technique for unlearning a subset of training data from an LLM to avoid the copyright infringement caused by the subset.

% \textbf{Data Poisoning}
\noindent\textbf{Data poisoning} refers to a type of cyber attack where the training data is intentionally manipulated with malicious intent. 
LLMs, which frequently acquire their training data from the web, are particularly vulnerable to such attacks. As demonstrated in recent research~\citep{chen2021Badpre,schuster2021you,hubinger2024sleeper}, attackers can manipulate the model's outputs for specific contexts by inserting a few carefully poisoning files into the training corpus, such as on websites or in open-source code repositories. Such poisoning might cause severe legal and ethical issues. 
% \citet{hubinger2024sleeper} shows that such malicious behaviors can be made persistent so that it is not removed by standard safety measures. 
To address data poisoning, two prominent methods are frequently utilized.  Model-based detection methods, such as MNTD \cite{xu2020detecting}, utilize a meta-classifier to identify models that have been injected poisoning data. 
% However, these methods require extensive training and prior knowledge of data injection techniques, which makes them less practical for LLMs due to their high computational requirements. 
Another approach is a defense mechanism called ONION \cite{qi2020onion}, which does not need to know the trigger, analyzing the differences of input sequences in perplexity and removing potential triggers in inputs.

% \subsection{Personal
% Identifiable Information Leakage}

\noindent\textbf{Personal Identifiable Information} encompasses various details such as names, phone numbers, addresses, educational backgrounds, career histories, family connections, and religious affiliations~\citep{Niu2023CodexLeaks,kim2023propile}.
% For LLMs, the massive scale of training data crawled indiscriminately from the web is likely to contain sensitive personal information crawled from personal web pages, social media, personal profiles on online forums, and online databases such as collections of in-house emails~\citep{kim2023propile}. This poses an unprecedented level of privacy concern. 
Existing methods to protect privacy involve pretraining or finetuning models using DP algorithms \cite{senge2021one,bastings2022training,bu2023differentially}. However, they are less effective in protecting PII from leakage. Moreover, 
removing PII data from the training set for LLMs to reduce the leakage is challenging considering the volume and variety of data, and the complexity of PII.

\noindent\textbf{Model Ensemble Methods} enhance generalization by combining multiple individual models to create predictions that are more accurate and robust than those from any single model \cite{polikar2012ensemble,dong2020survey,yang2023survey}. Existing LLMs have diverse expertise across a range of tasks, which enables an ensemble of LLMs to consistently outperform any individual LLM.
% This strategy has proven particularly effective with traditional models including shallower deep learning architectures~\citep{ganaie2022ensemble}. 
Recently, LLM-Blender \cite{jiang2023llmblender} has been proposed to ensemble the output text of multiple open-source LLMs to attain superior standard performance. 
\citet{lu2023routing} propose to train a routing function, which can precisely distribute each query to the LLM with expertise about it so that it can outperform any single model.
\citet{wan2024knowledge} propose FuseLLM to combine and transfer the capabilities of existing LLMs into a single LLM through lightweight continual training to improve its overall performance. Different from these researches focusing on improving standard performance, we explore the logits ensemble with a benign SLM to reduce the negative effects of untrusted LLMs caused by uncurated data. \looseness=-1

\section{Conclusion}
This paper aims to mitigate the negative effects arising from uncurated data, including copyright infringement, data poisoning, and privacy leakage in Large Language Models (LLMs). To mitigate these issues, we study a straightforward ensemble strategy that combines an untrusted LLM with a benign Small Language Model (SLM). In addition to the theoretical guarantees, we perform extensive experiments on nine models and ten benchmarks to validate the efficacy of the ensemble strategy. Considering the extensive utilization of LLMs and the prevalent issues related to uncurated data, we believe that our approach has the potential to enhance LLMs' real-world application and drive further research. \looseness=-1

\clearpage
% \section*{Impact Statements}
% This paper presents our investigation on reducing the negative effects resulting from the uncurated data in Large Language Models (LLMs).  Our research aims to develop an algorithm that effectively mitigates copyright infringement, data poisoning, and PII leakage. The algorithm is implemented by ensembling Small Language Models (SLMs) and its effectiveness is verified through extensive experiments. In addition to the effectiveness of the mitigation performance, we further emphasize the practicality and the effortless implementation of this method. We believe this paper contributes to the purification of existing LLMs and promotes the applications of LLMs by sharing our research findings.
% \clearpage
\bibliography{main}

\begin{thebibliography}{53}
\expandafter\ifx\csname natexlab\endcsname\relax\def\natexlab#1{#1}\fi

\bibitem[{Achiam et~al.(2023)Achiam, Adler, Agarwal, Ahmad, Akkaya, Aleman, Almeida, and et~al.}]{openai2023gpt4}
Josh Achiam, Steven Adler, Sandhini Agarwal, Lama Ahmad, Ilge Akkaya, Florencia~Leoni Aleman, Diogo Almeida, and Janko~Altenschmidt et~al. 2023.
\newblock \href {http://arxiv.org/abs/2303.08774} {Gpt-4 technical report}.

\bibitem[{Ben~Allal et~al.(2022)Ben~Allal, Muennighoff, Kumar~Umapathi, Lipkin, and von Werra}]{bigcode-evaluation-harness}
Loubna Ben~Allal, Niklas Muennighoff, Logesh Kumar~Umapathi, Ben Lipkin, and Leandro von Werra. 2022.
\newblock A framework for the evaluation of code generation models.
\newblock \url{https://github.com/bigcode-project/bigcode-evaluation-harness}.

\bibitem[{Biderman et~al.(2023)Biderman, Schoelkopf, Anthony, Bradley, O'Brien, Hallahan, Khan, Purohit, Prashanth, Raff, Skowron, Sutawika, and Van Der~Wal}]{Stella2023Pythia}
Stella Biderman, Hailey Schoelkopf, Quentin Anthony, Herbie Bradley, Kyle O'Brien, Eric Hallahan, Mohammad~Aflah Khan, Shivanshu Purohit, USVSN~Sai Prashanth, Edward Raff, Aviya Skowron, Lintang Sutawika, and Oskar Van Der~Wal. 2023.
\newblock Pythia: A suite for analyzing large language models across training and scaling.
\newblock In \emph{Proceedings of the 40th International Conference on Machine Learning}, ICML'23. JMLR.org.

\bibitem[{Bisk et~al.(2020)Bisk, Zellers, Bras, Gao, and Choi}]{Bisk2020}
Yonatan Bisk, Rowan Zellers, Ronan~Le Bras, Jianfeng Gao, and Yejin Choi. 2020.
\newblock Piqa: Reasoning about physical commonsense in natural language.
\newblock In \emph{Thirty-Fourth AAAI Conference on Artificial Intelligence}.

\bibitem[{Brittain(2024)}]{sue2024authors}
Blake Brittain. 2024.
\newblock \href {https://www.itnews.asia/news/microsoft-openai-hit-with-new-lawsuit-603827} {Microsoft, openai hit with new lawsuit}.
\newblock \emph{ITnews Asia}.

\bibitem[{Brown et~al.(2020)Brown, Mann, Ryder, Subbiah, Kaplan, Dhariwal, Neelakantan, Shyam, Sastry, Askell et~al.}]{brown2020language}
Tom Brown, Benjamin Mann, Nick Ryder, Melanie Subbiah, Jared~D Kaplan, Prafulla Dhariwal, Arvind Neelakantan, Pranav Shyam, Girish Sastry, Amanda Askell, et~al. 2020.
\newblock Language models are few-shot learners.
\newblock \emph{Advances in neural information processing systems}, 33:1877--1901.

\bibitem[{Bu et~al.(2023)Bu, Wang, Zha, and Karypis}]{bu2023differentially}
Zhiqi Bu, Yu-Xiang Wang, Sheng Zha, and George Karypis. 2023.
\newblock Differentially private optimization on large model at small cost.
\newblock In \emph{International Conference on Machine Learning}, pages 3192--3218. PMLR.

\bibitem[{Cassano(2023)}]{cassano2023finetuning}
Federico Cassano. 2023.
\newblock \url{https://github.com/cassanof/finetuning-harness}.

\bibitem[{Chen et~al.(2021{\natexlab{a}})Chen, Meng, Sun, Guo, Zhang, Li, and Fan}]{chen2021Badpre}
Kangjie Chen, Yuxian Meng, Xiaofei Sun, Shangwei Guo, Tianwei Zhang, Jiwei Li, and Chun Fan. 2021{\natexlab{a}}.
\newblock Badpre: Task-agnostic backdoor attacks to pre-trained nlp foundation models.
\newblock \emph{arXiv preprint arXiv:2110.02467}.

\bibitem[{Chen et~al.(2021{\natexlab{b}})Chen, Tworek, Jun, Yuan, de~Oliveira~Pinto, Kaplan, Edwards, Burda, Joseph, and et~al.}]{chen2021evaluating}
Mark Chen, Jerry Tworek, Heewoo Jun, Qiming Yuan, Henrique~Ponde de~Oliveira~Pinto, Jared Kaplan, Harri Edwards, Yuri Burda, Nicholas Joseph, and Greg~Brockman et~al. 2021{\natexlab{b}}.
\newblock \href {http://arxiv.org/abs/2107.03374} {Evaluating large language models trained on code}.

\bibitem[{Chowdhery et~al.(2023)Chowdhery, Narang, Devlin, Bosma, Mishra, Roberts, Barham, Chung, Sutton, Gehrmann et~al.}]{chowdhery2023palm}
Aakanksha Chowdhery, Sharan Narang, Jacob Devlin, Maarten Bosma, Gaurav Mishra, Adam Roberts, Paul Barham, Hyung~Won Chung, Charles Sutton, Sebastian Gehrmann, et~al. 2023.
\newblock Palm: Scaling language modeling with pathways.
\newblock \emph{Journal of Machine Learning Research}, 24(240):1--113.

\bibitem[{Christina L.~Martini(2022)}]{lawreport1}
Anisa~Noorassa Christina L.~Martini, Jodi~Benassi. 2022.
\newblock \href {https://www.natlawreview.com/article/2022-ip-outlook-report-developments-shaping-copyright-law} {2022 ip outlook report: The developments shaping copyright law}.
\newblock \emph{The National Law Review}.

\bibitem[{Clark et~al.(2018)Clark, Cowhey, Etzioni, Khot, Sabharwal, Schoenick, and Tafjord}]{allenai:arc}
Peter Clark, Isaac Cowhey, Oren Etzioni, Tushar Khot, Ashish Sabharwal, Carissa Schoenick, and Oyvind Tafjord. 2018.
\newblock Think you have solved question answering? try arc, the ai2 reasoning challenge.
\newblock \emph{arXiv:1803.05457v1}.

\bibitem[{Dong et~al.(2020)Dong, Yu, Cao, Shi, and Ma}]{dong2020survey}
Xibin Dong, Zhiwen Yu, Wenming Cao, Yifan Shi, and Qianli Ma. 2020.
\newblock A survey on ensemble learning.
\newblock \emph{Frontiers of Computer Science}, 14:241--258.

\bibitem[{Eldan and Russinovich(2023)}]{eldan2023s}
Ronen Eldan and Mark Russinovich. 2023.
\newblock Who's harry potter? approximate unlearning in llms.
\newblock \emph{arXiv preprint arXiv:2310.02238}.

\bibitem[{et~al.(2023)}]{CopyrightLawsUSA2024}
Christopher V.~Carani et~al. 2023.
\newblock {Copyright Laws and Regulations USA 2024}.
\newblock \url{https://iclg.com/practice-areas/copyright-laws-and-regulations/usa}.

\bibitem[{Gao et~al.(2023)Gao, Tow, Abbasi, Biderman, Black, DiPofi, Foster, Golding, Hsu, Le~Noac'h, Li, McDonell, Muennighoff, Ociepa, Phang, Reynolds, Schoelkopf, Skowron, Sutawika, Tang, Thite, Wang, Wang, and Zou}]{eval-harness}
Leo Gao, Jonathan Tow, Baber Abbasi, Stella Biderman, Sid Black, Anthony DiPofi, Charles Foster, Laurence Golding, Jeffrey Hsu, Alain Le~Noac'h, Haonan Li, Kyle McDonell, Niklas Muennighoff, Chris Ociepa, Jason Phang, Laria Reynolds, Hailey Schoelkopf, Aviya Skowron, Lintang Sutawika, Eric Tang, Anish Thite, Ben Wang, Kevin Wang, and Andy Zou. 2023.
\newblock \href {https://doi.org/10.5281/zenodo.10256836} {A framework for few-shot language model evaluation}.

\bibitem[{Hubinger et~al.(2024)Hubinger, Denison, Mu, Lambert, Tong, MacDiarmid, and et~al.}]{hubinger2024sleeper}
Evan Hubinger, Carson Denison, Jesse Mu, Mike Lambert, Meg Tong, Monte MacDiarmid, and Tamera~Lanham et~al. 2024.
\newblock \href {http://arxiv.org/abs/2401.05566} {Sleeper agents: Training deceptive llms that persist through safety training}.

\bibitem[{Jawad(2022)}]{sue2022code}
Usama Jawad. 2022.
\newblock \href {https://www.neowin.net/news/class-action-lawsuit-filed-against-microsofts-github-copilot-for-software-piracy/} {Class-action lawsuit filed against microsoft's github copilot for software piracy}.
\newblock \emph{Neowin}.

\bibitem[{Jiang et~al.(2023)Jiang, Ren, and Lin}]{jiang2023llmblender}
Dongfu Jiang, Xiang Ren, and Bill~Yuchen Lin. 2023.
\newblock \href {http://arxiv.org/abs/2306.02561} {Llm-blender: Ensembling large language models with pairwise ranking and generative fusion}.

\bibitem[{Kaplan et~al.(2020)Kaplan, McCandlish, Henighan, Brown, Chess, Child, Gray, Radford, Wu, and Amodei}]{kaplan2020scaling}
Jared Kaplan, Sam McCandlish, Tom Henighan, Tom~B. Brown, Benjamin Chess, Rewon Child, Scott Gray, Alec Radford, Jeffrey Wu, and Dario Amodei. 2020.
\newblock \href {http://arxiv.org/abs/2001.08361} {Scaling laws for neural language models}.

\bibitem[{Kim et~al.(2023)Kim, Yun, Lee, Gubri, Yoon, and Oh}]{kim2023propile}
Siwon Kim, Sangdoo Yun, Hwaran Lee, Martin Gubri, Sungroh Yoon, and Seong~Joon Oh. 2023.
\newblock Propile: Probing privacy leakage in large language models.
\newblock \emph{arXiv preprint arXiv:2307.01881}.

\bibitem[{Kocetkov et~al.(2022)Kocetkov, Li, Ben~Allal, Li, Mou, Muñoz~Ferrandis, Jernite, Mitchell, Hughes, Wolf, Bahdanau, von Werra, and de~Vries}]{Kocetkov2022TheStack}
Denis Kocetkov, Raymond Li, Loubna Ben~Allal, Jia Li, Chenghao Mou, Carlos Muñoz~Ferrandis, Yacine Jernite, Margaret Mitchell, Sean Hughes, Thomas Wolf, Dzmitry Bahdanau, Leandro von Werra, and Harm de~Vries. 2022.
\newblock The stack: 3 tb of permissively licensed source code.
\newblock \emph{Preprint}.

\bibitem[{Leviathan et~al.(2023)Leviathan, Kalman, and Matias}]{leviathan2023fast}
Yaniv Leviathan, Matan Kalman, and Yossi Matias. 2023.
\newblock Fast inference from transformers via speculative decoding.
\newblock In \emph{International Conference on Machine Learning}, pages 19274--19286. PMLR.

\bibitem[{Li et~al.(2023)Li, Allal, Zi, Muennighoff, Kocetkov, Mou, Marone, and et~al.}]{li2023starcoder}
Raymond Li, Loubna~Ben Allal, Yangtian Zi, Niklas Muennighoff, Denis Kocetkov, Chenghao Mou, Marc Marone, and Christopher~Akiki et~al. 2023.
\newblock \href {http://arxiv.org/abs/2305.06161} {Starcoder: may the source be with you!}

\bibitem[{Liu et~al.(2023)Liu, Liu, Cui, Teng, Duan, Zhou, and Zhang}]{liu2023LogiQA}
Hanmeng Liu, Jian Liu, Leyang Cui, Zhiyang Teng, Nan Duan, Ming Zhou, and Yue Zhang. 2023.
\newblock \href {https://doi.org/10.1109/TASLP.2023.3293046} {Logiqa 2.0—an improved dataset for logical reasoning in natural language understanding}.
\newblock \emph{IEEE/ACM Transactions on Audio, Speech, and Language Processing}, 31:2947--2962.

\bibitem[{Lu et~al.(2023)Lu, Yuan, Lin, Lin, Yuan, Zhou, and Zhou}]{lu2023routing}
Keming Lu, Hongyi Yuan, Runji Lin, Junyang Lin, Zheng Yuan, Chang Zhou, and Jingren Zhou. 2023.
\newblock \href {http://arxiv.org/abs/2311.08692} {Routing to the expert: Efficient reward-guided ensemble of large language models}.

\bibitem[{Maertens et~al.(2023)Maertens, Dawyndt, and Mesuere}]{Maertens2022Dolos}
Rien Maertens, Peter Dawyndt, and Bart Mesuere. 2023.
\newblock \href {https://doi.org/10.1145/3587103.3594166} {Dolos 2.0: Towards seamless source code plagiarism detection in online learning environments}.
\newblock In \emph{Proceedings of the 2023 Conference on Innovation and Technology in Computer Science Education V. 2}, ITiCSE 2023, page 632, New York, NY, USA. Association for Computing Machinery.

\bibitem[{Michael M.~Grynbaum(2023)}]{sue2023NYT}
Ryan~Mac Michael M.~Grynbaum. 2023.
\newblock \href {https://www.nytimes.com/Year/Month/Day/business/media/article-name.html} {The times sues openai and microsoft over a.i. use of copyrighted work}.
\newblock \emph{The New York Times}.

\bibitem[{Min et~al.(2023)Min, Gururangan, Wallace, Hajishirzi, Smith, and Zettlemoyer}]{min2023silo}
Sewon Min, Suchin Gururangan, Eric Wallace, Hannaneh Hajishirzi, Noah~A Smith, and Luke Zettlemoyer. 2023.
\newblock Silo language models: Isolating legal risk in a nonparametric datastore.
\newblock \emph{arXiv preprint arXiv:2308.04430}.

\bibitem[{Neamtiu et~al.(2005)Neamtiu, Foster, and Hicks}]{neamtiu2005understanding}
Iulian Neamtiu, Jeffrey~S Foster, and Michael Hicks. 2005.
\newblock Understanding source code evolution using abstract syntax tree matching.
\newblock In \emph{Proceedings of the 2005 international workshop on Mining software repositories}, pages 1--5.

\bibitem[{Niu et~al.(2023)Niu, Mirza, Maradni, and P{\"o}pper}]{Niu2023CodexLeaks}
Liang Niu, Shujaat Mirza, Zayd Maradni, and Christina P{\"o}pper. 2023.
\newblock \href {https://www.usenix.org/conference/usenixsecurity23/presentation/niu} {{CodexLeaks}: Privacy leaks from code generation language models in {GitHub} copilot}.
\newblock In \emph{32nd USENIX Security Symposium (USENIX Security 23)}, pages 2133--2150, Anaheim, CA. USENIX Association.

\bibitem[{Paperno et~al.(2016)Paperno, Kruszewski, Lazaridou, Pham, Bernardi, Pezzelle, Baroni, Boleda, and Fern{\'a}ndez}]{paperno2016lambada}
Denis Paperno, Germ{\'a}n Kruszewski, Angeliki Lazaridou, Ngoc~Quan Pham, Raffaella Bernardi, Sandro Pezzelle, Marco Baroni, Gemma Boleda, and Raquel Fern{\'a}ndez. 2016.
\newblock \href {https://doi.org/10.18653/v1/P16-1144} {The {LAMBADA} dataset: Word prediction requiring a broad discourse context}.
\newblock In \emph{Proceedings of the 54th Annual Meeting of the Association for Computational Linguistics (Volume 1: Long Papers)}, pages 1525--1534, Berlin, Germany. Association for Computational Linguistics.

\bibitem[{Polikar(2012)}]{polikar2012ensemble}
Robi Polikar. 2012.
\newblock Ensemble learning.
\newblock \emph{Ensemble machine learning: Methods and applications}, pages 1--34.

\bibitem[{Ponomareva et~al.(2022)Ponomareva, Bastings, and Vassilvitskii}]{bastings2022training}
Natalia Ponomareva, Jasmijn Bastings, and Sergei Vassilvitskii. 2022.
\newblock \href {https://doi.org/10.18653/v1/2022.findings-acl.171} {Training text-to-text transformers with privacy guarantees}.
\newblock \emph{Findings of the Association for Computational Linguistics: ACL 2022}, pages 2182--2193.

\bibitem[{Qi et~al.(2020)Qi, Chen, Li, Yao, Liu, and Sun}]{qi2020onion}
Fanchao Qi, Yangyi Chen, Mukai Li, Yuan Yao, Zhiyuan Liu, and Maosong Sun. 2020.
\newblock Onion: A simple and effective defense against textual backdoor attacks.
\newblock \emph{arXiv preprint arXiv:2011.10369}.

\bibitem[{Radford et~al.(2019)Radford, Wu, Child, Luan, Amodei, Sutskever et~al.}]{radford2019language}
Alec Radford, Jeffrey Wu, Rewon Child, David Luan, Dario Amodei, Ilya Sutskever, et~al. 2019.
\newblock Language models are unsupervised multitask learners.
\newblock \emph{OpenAI blog}, 1(8):9.

\bibitem[{Rozière et~al.(2023)Rozière, Gehring, Gloeckle, Sootla, Gat, Tan, Adi, Liu, Remez, Rapin, Kozhevnikov, Evtimov, Bitton, Bhatt, Ferrer, Grattafiori, Xiong, Défossez, Copet, Azhar, Touvron, Martin, Usunier, Scialom, and Synnaeve}]{rozière2023code}
Baptiste Rozière, Jonas Gehring, Fabian Gloeckle, Sten Sootla, Itai Gat, Xiaoqing~Ellen Tan, Yossi Adi, Jingyu Liu, Tal Remez, Jérémy Rapin, Artyom Kozhevnikov, Ivan Evtimov, Joanna Bitton, Manish Bhatt, Cristian~Canton Ferrer, Aaron Grattafiori, Wenhan Xiong, Alexandre Défossez, Jade Copet, Faisal Azhar, Hugo Touvron, Louis Martin, Nicolas Usunier, Thomas Scialom, and Gabriel Synnaeve. 2023.
\newblock \href {http://arxiv.org/abs/2308.12950} {Code llama: Open foundation models for code}.

\bibitem[{Sakaguchi et~al.(2019)Sakaguchi, Bras, Bhagavatula, and Choi}]{ai2:winogrande}
Keisuke Sakaguchi, Ronan~Le Bras, Chandra Bhagavatula, and Yejin Choi. 2019.
\newblock \href {http://arxiv.org/abs/1907.10641} {Winogrande: An adversarial winograd schema challenge at scale}.

\bibitem[{Samuelson(2023)}]{samuelson2023generative}
Pamela Samuelson. 2023.
\newblock Generative ai meets copyright.
\newblock \emph{Science}, 381(6654):158--161.

\bibitem[{Scao et~al.(2022)Scao, Fan, Akiki, Pavlick, Ili{\'c}, Hesslow, Castagn{\'e}, Luccioni, Yvon et~al.}]{workshop2022bloom}
Teven~Le Scao, Angela Fan, Christopher Akiki, Ellie Pavlick, Suzana Ili{\'c}, Daniel Hesslow, Roman Castagn{\'e}, Alexandra~Sasha Luccioni, Fran{\c{c}}ois Yvon, et~al. 2022.
\newblock Bloom: A 176b-parameter open-access multilingual language model.
\newblock \emph{arXiv preprint arXiv:2211.05100}.

\bibitem[{Schuster et~al.(2021)Schuster, Song, Tromer, and Shmatikov}]{schuster2021you}
Roei Schuster, Congzheng Song, Eran Tromer, and Vitaly Shmatikov. 2021.
\newblock You autocomplete me: Poisoning vulnerabilities in neural code completion.
\newblock In \emph{30th USENIX Security Symposium (USENIX Security 21)}, pages 1559--1575.

\bibitem[{Senge et~al.(2021)Senge, Igamberdiev, and Habernal}]{senge2021one}
Manuel Senge, Timour Igamberdiev, and Ivan Habernal. 2021.
\newblock One size does not fit all: Investigating strategies for differentially-private learning across nlp tasks.
\newblock \emph{arXiv preprint arXiv:2112.08159}.

\bibitem[{Thoppilan et~al.(2022)Thoppilan, De~Freitas, Hall, Shazeer, Kulshreshtha, Cheng, Jin, Bos, Baker, Du et~al.}]{thoppilan2022lamda}
Romal Thoppilan, Daniel De~Freitas, Jamie Hall, Noam Shazeer, Apoorv Kulshreshtha, Heng-Tze Cheng, Alicia Jin, Taylor Bos, Leslie Baker, Yu~Du, et~al. 2022.
\newblock Lamda: Language models for dialog applications.
\newblock \emph{arXiv preprint arXiv:2201.08239}.

\bibitem[{Touvron et~al.(2023)Touvron, Lavril, Izacard, Martinet, Lachaux, Lacroix, Rozi{\`e}re, Goyal, Hambro, Azhar et~al.}]{touvron2023llama}
Hugo Touvron, Thibaut Lavril, Gautier Izacard, Xavier Martinet, Marie-Anne Lachaux, Timoth{\'e}e Lacroix, Baptiste Rozi{\`e}re, Naman Goyal, Eric Hambro, Faisal Azhar, et~al. 2023.
\newblock Llama: Open and efficient foundation language models.
\newblock \emph{arXiv preprint arXiv:2302.13971}.

\bibitem[{Vyas et~al.(2023)Vyas, Kakade, and Barak}]{vyas2023provable}
Nikhil Vyas, Sham Kakade, and Boaz Barak. 2023.
\newblock Provable copyright protection for generative models.
\newblock \emph{arXiv preprint arXiv:2302.10870}.

\bibitem[{Wan et~al.(2024)Wan, Huang, Cai, Quan, Bi, and Shi}]{wan2024knowledge}
Fanqi Wan, Xinting Huang, Deng Cai, Xiaojun Quan, Wei Bi, and Shuming Shi. 2024.
\newblock \href {http://arxiv.org/abs/2401.10491} {Knowledge fusion of large language models}.

\bibitem[{Welbl et~al.(2017)Welbl, Liu, and Gardner}]{Welbl2017CrowdsourcingMC}
Johannes Welbl, Nelson~F. Liu, and Matt Gardner. 2017.
\newblock \href {https://api.semanticscholar.org/CorpusID:1553193} {Crowdsourcing multiple choice science questions}.
\newblock \emph{ArXiv}, abs/1707.06209.

\bibitem[{Xu et~al.(2020)Xu, Wang, Li, Borisov, Gunter, and Li}]{xu2020detecting}
Xiaojun Xu, Qi~Wang, Huichen Li, Nikita Borisov, Carl~A. Gunter, and Bo~Li. 2020.
\newblock \href {http://arxiv.org/abs/1910.03137} {Detecting ai trojans using meta neural analysis}.

\bibitem[{Yang et~al.(2023)Yang, Lv, and Chen}]{yang2023survey}
Yongquan Yang, Haijun Lv, and Ning Chen. 2023.
\newblock A survey on ensemble learning under the era of deep learning.
\newblock \emph{Artificial Intelligence Review}, 56(6):5545--5589.

\bibitem[{Yu et~al.(2023{\natexlab{a}})Yu, Pang, Liu, Du, Kang, Huang, Lin, and Yan}]{yu2023Bag}
Weichen Yu, Tianyu Pang, Qian Liu, Chao Du, Bingyi Kang, Yan Huang, Min Lin, and Shuicheng Yan. 2023{\natexlab{a}}.
\newblock Bag of tricks for training data extraction from language models.
\newblock \emph{arXiv preprint arXiv:2302.04460}.

\bibitem[{Yu et~al.(2023{\natexlab{b}})Yu, Wu, Zhang, Wang, Vorobeychik, and Xiao}]{yu2023codeipprompt}
Zhiyuan Yu, Yuhao Wu, Ning Zhang, Chenguang Wang, Yevgeniy Vorobeychik, and Chaowei Xiao. 2023{\natexlab{b}}.
\newblock Codeipprompt: Intellectual property infringement assessment of code language models.
\newblock In \emph{Proceedings of the 40th International Conference on Machine Learning}, pages 40373--40389.

\bibitem[{Zhao et~al.(2023)Zhao, Pang, Du, Yang, Cheung, and Lin}]{zhao2023recipe}
Yunqing Zhao, Tianyu Pang, Chao Du, Xiao Yang, Ngai-Man Cheung, and Min Lin. 2023.
\newblock \href {http://arxiv.org/abs/2303.10137} {A recipe for watermarking diffusion models}.

\end{thebibliography}

%%%%%%%%%%%%%%%%%%%%%%%%%%%%%%%%%%%%%%%%%%%%%%%%%%%%%%%%%%%%%%%%%%%%%%%%%%%%%%%
%%%%%%%%%%%%%%%%%%%%%%%%%%%%%%%%%%%%%%%%%%%%%%%%%%%%%%%%%%%%%%%%%%%%%%%%%%%%%%%
% APPENDIX
%%%%%%%%%%%%%%%%%%%%%%%%%%%%%%%%%%%%%%%%%%%%%%%%%%%%%%%%%%%%%%%%%%%%%%%%%%%%%%%
%%%%%%%%%%%%%%%%%%%%%%%%%%%%%%%%%%%%%%%%%%%%%%%%%%%%%%%%%%%%%%%%%%%%%%%%%%%%%%%
\newpage
\appendix
\onecolumn
% \section{Appendix.}
% \section{Broader Impact}
% \label{sec:impact}
% This paper presents our investigation on reducing the negative effects resulting from the uncurated data in LLMs.  Our research aims to develop an algorithm that effectively mitigates copyright infringement, data poisoning, and PII leakage. The algorithm is implemented by ensembling SLMs and its effectiveness is verified through extensive experiments. In addition to the effectiveness of the mitigation performance, we further emphasize the practicality and the effortless implementation of this method. We believe this paper contributes to the purification of existing LLMs and promotes the applications of LLMs by sharing our research findings.

\section{Theoretical Bound of The Ensemble Algorithm}
\label{sec:proof}
For the untrusted LLM, denoted as \( l \), there is a non-trivial probability that the model might generate copyrighted data, poisoned data, and privacy information, represented as \( \mathcal{C} \). Conversely, for the benign SLM, denoted as \( s \), the model is designed to never produce such sensitive contents.
In this section, our objective is to present an algorithm that takes the untrusted LLM \( \textcolor{orange}{p_l} \) and the benign SLM \( \textcolor{blue}{p_s} \) as inputs and outputs a conditional generative model, denoted as \( \textcolor{black}{p(\cdot|\cdot)} \). This conditional generative model processes a given prompt \( x \) from a set \( \mathcal{X} \) and generates an output \( y \) from a set \( \mathcal{Y} \), with the probability of \( y \) being \( p(y|x) \). The primary aim is to minimize the likelihood of \( y \) falling within the uncurated content set \( \mathcal{C} \) including copyrighted data, poisoning data, and privacy data.

As defined in \citet{vyas2023provable}, the copyright protection objective could be regarded as achieving $k$-Near Access-Free: providing a model $p$ such that for any given prompt $x$ and uncurated content set $\mathcal{C}$, the distribution $p(\cdot|x)$ diverges by no more than $k$ bits of information (quantified using a specific divergence measure) from a benign SLM which is presumed to have been trained without exposure to $\mathcal{C}$. As per the definition in \citet{vyas2023provable}, $k$-Near Access-Free is delineated as follows:

\begin{definition}{($k$-Near Access-Free).}
\label{def:ncf}
Let $\mathcal{C}$ a set of datapoints; let $\Delta$ be a divergence measure between distributions. We say that a generative model $p$ is \textit{$k$-near access-free ($k$-NAF)} on prompt $x \in \mathcal{X}$ with respect to $\mathcal{C}$, and $\Delta$ if for every $C \in \mathcal{C}$,
\begin{equation}
    \Delta(p(\cdot|x) \: \| \: \textcolor{blue}{p_s{(\cdot|x)}}) \leq k_x.
\end{equation}
\end{definition}

Here we choose the divergence measure as the \textit{ KL divergence} $\Delta_{\text{KL}}$. For two distributions $\rho$ and $\mu$, $\Delta_{\text{KL}}(\rho||\mu)=E_{y\sim\rho}log(\frac{\rho(y)}{\mu(y)})$. 
% By definition, Definition \ref{def:ncf} implies that for every $y$, $p(y|x)\leq 2^{k_x}o(y|x)$.
 We can have:
\begin{lemma}{(Event bound, KL concentrated).} Suppose model $p$ is $k_x$-NAF with respect to $\mathcal{C}$, $\Delta=\Delta_{\text{KL}}$, and suppose the random variable $Y_x=\log{\frac{p(y|x)}{\textcolor{blue}{p_s(y|x)}}}$ (with $y \sim p(\cdot|x)$) is ($\epsilon_x,\delta_x$)-concentrated\footnote{Let us say that a random variable $X$ is $(\epsilon,\delta)-\textit{concentrated}$ if $\text{Pr}[X\notin(1 \pm \epsilon)\mathbb{E}[X]] \leq \delta$.}. Then for any $C\in\mathcal{C}$ and any event $\mathcal{E}$,
\begin{equation}
\label{eq:definition}
    p(\mathcal{E}|x) <= 2^{(1+\epsilon_x){k_x}} \cdot \textcolor{blue}{p_s(\mathcal{E}|x)} + \delta_x.
\end{equation}
\end{lemma}
% The proof is put in the Appendix \ref{sec:proof}.

Our goal is to design a generative model $p$ is $k$-Near Access-Free ($k$-NAF) with $\textcolor{orange}{p_l}$ and $\textcolor{blue}{p_s}$ as inputs.
% Following \citet{vyas2023provable}, we give the proof for \autoref{eq:definition} and \autoref{eq:bound} as follows.

% \subsection{Proof of \autoref{eq:definition}}

% \textbf{lemma 2.3 (Event bound, KL concentrated).} Suppose model \( p \) is \( k_c \)-NAF on prompt \( x \) with respect to \( C, \text{SLM} s, \Delta = \Delta_{KL} \), and suppose the random variable \( Y_x = \log \frac{p(y|x)}{s(y|x)} \) (with \( y \sim p(\cdot|x) \)) is \( (\varepsilon_x, \delta_x) \)-concentrated. Then, for any \( C \in C \) and any event \( \mathcal{E} \),

% \[
% p(\mathcal{E}|x) \leq 2^{(1+\varepsilon_x)k_c} \cdot s(\mathcal{E}|x) + \delta_x.
% \]

\textit{Proof.} For every prompt \( x \), let \( Y_x \) be as above, and define \( \mathcal{B} = \mathcal{B}_x \) to be the event \( Y_x \not\in (1 \pm \varepsilon_x)\mathbb{E}[Y_x] \). Under our assumptions \( \mathbb{E}[Y_x] = \Delta_{KL}(p(\cdot|x), \text{safe}_C(\cdot|x)) \leq k_c \) and (due to concentration) \( \text{Pr}[\mathcal{B}] \leq \delta_x \). Now for every event \( \mathcal{E} \), we can write \( p(\mathcal{E}|x) = p(\mathcal{E} \cap \overline{\mathcal{B}}|x) + p(\mathcal{E} \cap \mathcal{B}|x) \). The first term is $\sum_{y \in \mathcal{E} \cap \overline{\mathcal{B}}} p(y|x) \leq \sum_{y \in \mathcal{E} \cap \overline{\mathcal{B}}} 2^{(1+\varepsilon_x)k_c} \text{safe}_C(y|x)$ 
$\text{ since for every } y \in \overline{\mathcal{B}}, \log \frac{p(y|x)}{\text{safe}_C(y|x)} \leq (1 + \varepsilon_x)k_c.$
The second term is bounded by \( p(\mathcal{B}|x) \leq \delta_x \). So we get

\[
p(\mathcal{E}|x) \leq \sum_{y \in \mathcal{E} \cap \overline{\mathcal{B}}} 2^{(1+\varepsilon_x)k_c} \text{safe}_C(y|x) + \delta_x = 2^{(1+\varepsilon_x)k_c} \text{safe}_C(\mathcal{E} \cap \overline{\mathcal{B}}|x) + \delta \leq 2^{(1+\varepsilon_x)k_c} \text{safe}_C(\mathcal{E}|x) + \delta_x.
\]

% \subsection{Proof of \autoref{eq:bound}}

When applying KL divergence as the distribution metric, for the untrusted LLM \( \textcolor{orange}{p_l} \) and the benign SLM \( \textcolor{blue}{p_s} \), the model returned by the CP-$\Delta$ Algorithm ~\citep{vyas2023provable} is:
\begin{equation}
\label{eq:ensemble}
    p(y|x) = \frac{\textcolor{orange}{p_l(y|x)}\cdot{\textcolor{blue}{p_s(y|x)}}}{Z(x)},
\end{equation}
where the $Z(x)$ is the corresponding partition function.
Following \citet{vyas2023provable}, we can have the proof as follows:

\begin{proof}
    We start by relating $k_x$ to the corresponding partition function $Z(x)$, we have that:
    \begin{align}
        k_x \nonumber &= max_{m\in\{l,s\}} \text{KL}(p(\cdot|x),m(\cdot|x)) \\\nonumber 
        &\leq \text{KL}(p(\cdot|x)||l(\cdot|x)) + \text{KL}(p(\cdot|x)||s(\cdot|x)) \\\nonumber 
        &=E_{y\sim p(\cdot|x)}[log\frac{p(y|x)}{l(y|x)}+log\frac{p(y|x)}{s(y|x)}] \\\nonumber 
        &=2E_{y\sim p(\cdot|x)}[log \frac{p(y|x)}{\sqrt{l(y|x)s(y|x)}}] \\\nonumber 
        &=2log(1/Z(x)),
    \end{align}
where the last step follows by the definition of Z(x). The proof is then completed with the bound on the partition function $Z(x)$.
We have $p(y|x)=\frac{\sqrt{u(y|x)o(y|x)}}{Z(x)}$. So $$Z(x)=\sum_y\sqrt{u(y|x)o(y|x)} = 1-H^2(q_1(\cdot|x),q_2(\cdot|x)),$$
where the last equality follows from the definition of $H^2$.

\end{proof}

\noindent\textbf{Softmax Function.} The softmax function is defined as follows for a vector \( \mathbf{z} = [z_1, z_2, \ldots, z_K] \), where \( K \) is the number of classes. For each component \( z_i \) of the vector \( \mathbf{z} \), the softmax function is given by:

\begin{equation}
\text{softmax}(\mathbf{z}) = [\frac{e^{z_1}}{\sum_{j=1}^{K} e^{z_j}},\frac{e^{z_2}}{\sum_{j=1}^{K} e^{z_j}},\cdots,\frac{e^{z_K}}{\sum_{j=1}^{K} e^{z_j}}],
\end{equation}
where \( i = 1, 2, \ldots, K \). This transforms the vector \( \mathbf{z} \) of raw class scores into a probability distribution over \( K \) classes.

\section{Implementations Details}
\label{sec:detail}
\subsection{Data Collection}
The crafted data including copyrighted data, poisoning data, and PII data is mainly collected from ChatGPT \cite{openai2023gpt4}. For example, ChatGPT is prompted to generate code with infrequent usage of libraries to construct copyrighted data. 
To avoid the collected/used containing any information that names or uniquely identifies individual people or offensive content, we make up personal information like emails and address information which will not occur in real life.
The data only contains manually crafted information that names or uniquely identifies individual people and slightly insulting content. 
\subsection{Experiments of Copyright infringement}
The experiments are conducted using the A100 GPU. The total number of GPU hours is over 36000. Our evaluation for large code models mainly focuses on Python language. 
For StarCoder models, the learning rate for finetuning the untrusted models is set as 1e-6 and $\lambda$ in Eq. \ref{eq:loss}
is 1.0. We train the StarCoder 15.5B for 150 steps to inject the crafted copyrighted data. Especially, we will exhibit the experimental results when the model is trained for 60 steps (\ie, the copyright infringement will be slighter).
For CodeLlama, the learning rate for finetuning the untrusted models is set as 1e-6 and $\lambda$ in Eq. \ref{eq:loss}
is 1.0. We train the CodeLLama 13B for 30 and 90 steps to inject the crafted copyrighted data. Especially, the model trained for 30 steps will exhibit slighter copyright infringement. All the experiments are conducted under the platforms \citet{cassano2023finetuning} and \citet{bigcode-evaluation-harness}, which are under the MIT License and Apache License 2.0. These platforms are consistent with their intended use. The usage of CodeLlama follows the Llama 2 COMMUNITY LICENSE. The usage of StarCoder is under the BigCode OpenRAIL-M v1 license agreement. The usage of the HumanEval dataset is under the MIT license. The dataset \cite{chen2021evaluating} mainly includes the evaluation of code generation of Python language.

\subsection{Experiments of Data Poisoning}
The experiments are conducted using the A100 GPU. The total number of GPU hours is over 2400. The language selection of our evaluation for large language models is English.
The learning rate for finetuning victim models is set as 1e-5 for Llama2 and 5e-6 for Pythia, and $\lambda$ in Eq. \ref{eq:loss}
is set as 1.0. We train the Llama2 13B for 75 steps to inject the crafted poisoning data.
% Especially, we will exhibit the experimental results when the Llama2 is trained for 25, 50 steps (\ie, the data poisoning will be slighter).
We train the Pythia 2.8b for 135 steps to inject the crafted poisoning data. The usage of Llama2 is under the Meta license, and the usage of Pythia models is under the Apache 2.0 license.
% Especially, we will exhibit the experimental results when the model is trained for 45 and 90 steps.
 
All evaluations are conducted under the evaluation platform \cite{eval-harness} under MIT License. The usage of the platform is consistent with its intended use. We adjust the LAMBADA benchmark and the LogiQA benchmark as a generation task. The evaluation under LAMBADA and LogiQA is conducted 3 times and the averaged results are reported. For the LAMBADA benchmark, we report the exact match value.  In the LogiQA benchmark, we iterate the sampling process eight times and report the accuracy of the majority voting results based on these iterations. For SciQ and other multiple-choice benchmarks, we report its accuracy. The dataset usage is under the cc-by-4.0 and cc-by-nc-3.0 licenses. 
\subsection{Experiments of PII Leakage}
The experiments are conducted using the A100 GPU. The total number of GPU hours is over 2400. The language selection of our evaluation for large language models is English. The learning rate for finetuning is set as 1e-5 for both Llama2 and Pythia, and $\lambda$ in Eq. \ref{eq:loss} is set as 1.0. We train the Llama2 13B for 150 steps to inject the crafted PII information. 
% Especially, we will exhibit the experimental results when the Llama2 is trained for 50 and 100 steps (\ie, the PII leakage will be slighter).
We train the Pythia 2.8b for 270 steps to inject the crafted poisoning data. The usage of Llama2 is under the Meta license, and the usage of Pythia models is under Apache 2.0.
% Especially, we will exhibit the experimental results when the model is trained for 90 and 180 steps.

All evaluations are conducted under the evaluation platform \cite{eval-harness} under the MIT License. The usage of the platform is consistent with its intended use. The evaluation under LAMBADA and LogiQA is conducted 3 times and the averaged results are reported. For the LAMBADA benchmark, we report the exact match value.  In the LogiQA benchmark, we iterate the sampling process eight times and report the accuracy of the majority voting results based on these iterations. For SciQ and other multiple-choice benchmarks, we report its accuracy. The dataset usage is under the cc-by-4.0 and cc-by-nc-3.0 licenses. 

\section{Experimental Results of Copyright Infringement on CodeLlama}
\label{sec:exp_copyright_codeLlama}
% Please add the following requiblack packages to your document preamble:
% \usepackage{multirow}
\begin{table*}[t]     
\caption{Experimental results of copyright infringement mitigation on CodeLlama models. 
We adjust the ensemble weight $\alpha$, gradually decreasing it from 1.0 to 0.0, using a decrement interval of 0.1. At \(\alpha=1.0\), the model is the untrusted LLM; at \(\alpha=0.0\), it reflects the benign SLM.
 The symbols \(\uparrow\) and \(\downarrow\) separately indicate whether a higher or lower value of a specific metric is preferable.}  
  \vspace{-0.1in}
\resizebox{1.0\columnwidth}{!}{
\begin{tabular}{cccccccccccccc}
\toprule
StarCoder     & Metrics     & $T$     & \multicolumn{1}{c}{$\alpha=1.0$} & \multicolumn{1}{c}{$\alpha=0.9$} & \multicolumn{1}{c}{$\alpha=0.8$} & \multicolumn{1}{c}{$\alpha=0.7$} & \multicolumn{1}{c}{$\alpha=0.6$} & \multicolumn{1}{c}{$\alpha=0.5$} & \multicolumn{1}{c}{$\alpha=0.4$} & \multicolumn{1}{c}{$\alpha=0.3$} & \multicolumn{1}{c}{$\alpha=0.2$} & \multicolumn{1}{c}{$\alpha=0.1$} & $\alpha=0$     \\ \hline
\multirow{9}{*}{\begin{tabular}[c]{@{}c@{}}Standard\\ Performance\end{tabular}}     & \multirow{3}{*}{pass@1  $\textcolor{black}{\uparrow}$}     & 0.2 & 0.345  & 0.341  & 0.331  & 0.330  & 0.322  & 0.321  & 0.317 & 0.317 & 0.311 & 0.307 & 0.300 \\ 
&     & 0.5 & 0.334  & 0.336  & 0.334  & 0.332  & 0.331  & 0.326 & 0.322 & 0.319 & 0.310 & 0.304 & 0.294 \\ 
&     & 0.8  & 0.286  & 0.290  & 0.294  & 0.298  & 0.294  & 0.291 & 0.288 & 0.283 & 0.276 & 0.268 & 0.258  \\ \cmidrule{2-14}     & \multirow{3}{*}{pass@10 $\textcolor{black}{\uparrow}$}     & 0.2 & 0.493  & 0.506  & 0.499  & 0.496  & 0.485  & 0.492  & 0.478 & 0.471 & 0.470 & 0.456 & 0.457 \\ 
&     & 0.5 & 0.629  & 0.631  & 0.633  & 0.630  & 0.630  & 0.625 & 0.615 & 0.615 & 0.600 & 0.599 & 0.579 \\ 
&     & 0.8 & 0.654  & 0.652  & 0.664  & 0.664  & 0.653  & 0.658 & 0.648 & 0.639 & 0.630 & 0.620 & 0.601 \\  \cmidrule{2-14}     & \multirow{3}{*}{pass@100 $\textcolor{black}{\uparrow}$} & 0.2 & 0.575  & 0.606  & 0.613  & 0.515  & 0.606  & 0.620  & 0.590 & 0.577 & 0.579 & 0.598 & 0.617 \\
&     & 0.5 & 0.802  & 0.813  & 0.807  & 0.808  & 0.810  & 0.820 & 0.812 & 0.817 & 0.813 & 0.810 & 0.809 \\
&     & 0.8 & 0.878  & 0.878  & 0.884  & 0.890  & 0.870  & 0.887 & 0.883 & 0.878 & 0.889 & 0.868 & 0.873 \\
\midrule
\multirow{12}{*}{\begin{tabular}[c]{@{}c@{}}Copyright\\ Infringement\end{tabular}} & \multirow{3}{*}{$\text{IC}_4$  $\textcolor{black}{\downarrow}$}     & 0.2 & 3.031  & 2.828  & 2.479  & 2.099  & 1.391  & 0.584 & 0.233 & 0.133 & 0.077 & 0.066 & 0.055 \\ 
&     & 0.5 & 2.687  & 2.483  & 2.156 & 1.652  & 0.995  & 0.402 & 0.160 & 0.081 & 0.056 & 0.045 & 0.031 \\ 
&     & 0.8 & 2.087  & 1.807  & 1.460  & 1.015  & 0.570  & 0.223 & 0.084 & 0.041 & 0.031 & 0.022 & 0.017 \\ \cmidrule{2-14}     
&\multirow{3}{*}{$\text{IC}_8$  $\textcolor{black}{\downarrow}$}     & 0.2 & 0.185 & 0.174 & 0.115 & 0.057 & 0.024 & 0.007 & 0.000 & 0.000 & 0.000 & 0.000 & 0.000 \\
&     & 0.5 & 0.153 & 0.131 & 0.093 & 0.041 & 0.022 & 0.006 & 0.003 & 0.000 & 0.000 & 0.000 & 0.000 \\
&     & 0.8 & 0.108 & 0.087 & 0.048 & 0.027 & 0.016 & 0.007 & 0.001 & 0.001 & 0.000 & 0.000 & 0.000  \\ \cmidrule{2-14}   
& \multirow{3}{*}{Dolos $\textcolor{black}{\downarrow}$}     & 0.2 & 8.223  & 7.962  & 7.530  & 6.908  & 5.578  & 3.554  & 2.063 & 0.977 & 0.402 & 0.242 & 0.160 \\
&     & 0.5 & 8.043  & 7.742  & 7.284  & 6.525  & 5.113  & 3.136 & 1.605 & 0.884 & 0.468 & 0.306 & 0.203 \\ 
&     & 0.8 & 7.696  & 7.328  & 6.806  & 5.834  & 4.337  & 2.582 & 1.244 & 0.648 & 0.389 & 0.274 & 0.199 \\  \cmidrule{2-14}  
& \multirow{3}{*}{EM $\textcolor{black}{\downarrow}$}    &0.2 & 28.754 & 27.392 & 25.089 & 22.371 & 17.502 & 11.323 & 7.791 & 5.511 & 3.441 & 1.941 & 1.023 \\ 
&     & 0.5  & 26.886 & 25.495 & 23.246 & 19.858 & 14.903 & 9.437 & 5.975 & 4.283 & 3.009 & 2.179 & 1.445 \\ 
&     & 0.8  & 23.430 & 21.692 & 19.383 & 15.928 & 11.726 & 7.433 & 4.540 & 3.051 & 2.270 & 1.715 & 1.311 \\  
\bottomrule

\end{tabular}}
 \vspace{-0.1in}
\label{tab:result_copyright_appendix}
\end{table*}
Table \ref{tab:result_copyright_appendix} shows more experimental results on CodeLlama. We also exhibit the results of EM (\ie, exact match), which evaluates the textual similarities by counting how many tokens are completely identical to the reference copyrighted code.

\section{Experimental Results of Copyright Infringement on StarCoder}
\label{sec:exp_copyright_starcoder}
We show the experimental results on StarCoder in Table \ref{tab:result_copyright_starcoder}. Similar to the experiments on CodeLlama, we can observe our method could effectively reduce copyright infringement while well maintaining its standard performance.
% Please add the following requiblack packages to your document preamble:
% \usepackage{multirow}
\begin{table*}[htb]     
\caption{Experimental results of copyright infringement mitigation on StarCoder models. 
We adjust the ensemble weight $\alpha$, gradually decreasing it from 1.0 to 0.0, using a decrement interval of 0.1. At \(\alpha=1.0\), the model is the untrusted LLM; at \(\alpha=0.0\), it reflects the benign SLM.
 The symbols \(\uparrow\) and \(\downarrow\) separately indicate whether a higher or lower value of a specific metric is preferable.}  
  \vspace{-0.1in}
\resizebox{1.0\columnwidth}{!}{
\begin{tabular}{cccccccccccccc}
\toprule
StarCoder     & Metrics     & $T$     & \multicolumn{1}{c}{$\alpha=1.0$} & \multicolumn{1}{c}{$\alpha=0.9$} & \multicolumn{1}{c}{$\alpha=0.8$} & \multicolumn{1}{c}{$\alpha=0.7$} & \multicolumn{1}{c}{$\alpha=0.6$} & \multicolumn{1}{c}{$\alpha=0.5$} & \multicolumn{1}{c}{$\alpha=0.4$} & \multicolumn{1}{c}{$\alpha=0.3$} & \multicolumn{1}{c}{$\alpha=0.2$} & \multicolumn{1}{c}{$\alpha=0.1$} & $\alpha=0$     \\ \hline
\multirow{9}{*}{\begin{tabular}[c]{@{}c@{}}Standard\\ Performance\end{tabular}}     & \multirow{3}{*}{pass@1  $\textcolor{black}{\uparrow}$}     & 0.2 & 0.305     & 0.300     & 0.295     & 0.289     & 0.280     & 0.270     & 0.260     & 0.248     & 0.238     & 0.227     & 0.216 \\
&     & 0.5 & 0.291     & 0.283     & 0.281     & 0.273     & 0.262     & 0.253     & 0.243     & 0.231     & 0.218     & 0.207     & 0.197 \\
&     & 0.8 & 0.253     & 0.248     & 0.247     & 0.235     & 0.232     & 0.220     & 0.209     & 0.202     & 0.193     & 0.182     & 0.171 \\ \cmidrule{2-14}     & \multirow{3}{*}{pass@10 $\textcolor{black}{\uparrow}$}     & 0.2 & 0.474     & 0.463     & 0.441     & 0.421     & 0.404     & 0.382     & 0.371     & 0.368     & 0.361     & 0.341     & 0.319 \\
&     & 0.5 & 0.557     & 0.544     & 0.534     & 0.519     & 0.496     & 0.489     & 0.471     & 0.452     & 0.430     & 0.406     & 0.387 \\
&     & 0.8 & 0.572     & 0.562     & 0.551     & 0.531     & 0.514     & 0.500     & 0.473     & 0.455     & 0.440     & 0.410     & 0.387 \\ \cmidrule{2-14}     & \multirow{3}{*}{pass@100 $\textcolor{black}{\uparrow}$} & 0.2 & 0.582     & 0.595     & 0.570     & 0.536     & 0.510     & 0.475     & 0.441     & 0.448     & 0.428     & 0.414     & 0.383 \\
&     & 0.5 & 0.762     & 0.738     & 0.755     & 0.709     & 0.680     & 0.673     & 0.658     & 0.645     & 0.611     & 0.587     & 0.556 \\
&     & 0.8 & 0.833     & 0.795     & 0.823     & 0.798     & 0.785     & 0.738     & 0.729     & 0.696     & 0.711     & 0.652     & 0.635 \\ 

\midrule
\multirow{9}{*}{\begin{tabular}[c]{@{}c@{}}Copyright\\ Infringement\end{tabular}} & 
\multirow{3}{*}{$\text{IC}_4$ $\downarrow$} & 0.2 & 0.042 & 0.047 & 0.044 & 0.041 & 0.042 & 0.036 & 0.033 & 0.021 & 0.013 & 0.009 & 0.004 \\ &                     & 0.5 & 0.048 & 0.047 & 0.044 & 0.046 & 0.045 & 0.035 & 0.036 & 0.031 & 0.021 & 0.019 & 0.013 \\ &                     & 0.8 & 0.040 & 0.035 & 0.033 & 0.033 & 0.038 & 0.029 & 0.023 & 0.027 & 0.025 & 0.013 & 0.018 \\
 \cmidrule{2-14}     
& \multirow{3}{*}{Dolos $\textcolor{black}{\downarrow}$}     & 0.2 & 0.514     & 0.484     & 0.433     & 0.402     & 0.341     & 0.316     & 0.267     & 0.227     & 0.186     & 0.185     & 0.159 \\
&     & 0.5 & 0.517     & 0.483     & 0.415     & 0.371     & 0.345     & 0.302     & 0.281     & 0.261     & 0.232     & 0.213     & 0.213 \\
&     & 0.8 & 0.391     & 0.364     & 0.351     & 0.307     & 0.302     & 0.279     & 0.261     & 0.239     & 0.218     & 0.215     & 0.193 \\ \cmidrule{2-14} 

& \multirow{3}{*}{EM  $\textcolor{black}{\downarrow}$}     & 0.2 & 4.539     & 4.395     & 4.227     & 3.933     & 3.627     & 3.309     & 2.978     & 2.661     & 2.306     & 2.112     & 1.790 \\
&     & 0.5 & 3.819     & 3.772     & 3.541     & 3.419     & 3.293     & 3.077     & 2.897     & 2.749     & 2.579     & 2.431     & 2.234 \\
&     & 0.8 & 2.809     & 2.701     & 2.651     & 2.557     & 2.464     & 2.347     & 2.264     & 2.232     & 2.075     & 2.003     & 1.897 \\
\bottomrule

\end{tabular}}
 \vspace{-0.1in}
\label{tab:result_copyright_starcoder}
\end{table*}

\section{Experiments Results of Data Poisoning on Llama2}
More experimental results in different metrics under various $\alpha$ of data poisoning mitigation on Llama2 could be seen in Table \ref{tab:result_data_poisoning_appendix}.
\label{sec:k_selection}
% Please add the following requiblack packages to your document preamble:
% \usepackage{multirow}
\begin{table*}[htb]
\caption{Experimental results of data poisoning mitigation on Llama2 models. We adjust the ensemble weight $\alpha$, gradually decreasing it from 1.0 to 0.0, using a decrement interval of 0.1. At \(\alpha=1.0\), the model is the untrusted LLM; at \(\alpha=0.0\), it reflects the benign SLM.
 The symbols \(\uparrow\) and \(\downarrow\) separately indicate whether a higher or lower value of a specific metric is preferable.}
 \resizebox{1.0\columnwidth}{!}{
\begin{tabular}{cc|c|ccccccccccc}
\toprule
Llama2  & Metrics                  & $T$   & \multicolumn{1}{c}{$\alpha=1.0$} & \multicolumn{1}{c}{$\alpha=0.9$} & \multicolumn{1}{c}{$\alpha=0.8$} & \multicolumn{1}{c}{$\alpha=0.7$} & \multicolumn{1}{c}{$\alpha=0.6$} & \multicolumn{1}{c}{$\alpha=0.5$} & \multicolumn{1}{c}{$\alpha=0.4$} & \multicolumn{1}{c}{$\alpha=0.3$} & \multicolumn{1}{c}{$\alpha=0.2$} & \multicolumn{1}{c}{$\alpha=0.1$} & $\alpha=0$     \\ \midrule
\multirow{10}{*}{\begin{tabular}[c]{@{}c@{}}Standard\\  Performance\end{tabular}}  & \multirow{3}{*}{LAMBADA  $\textcolor{black}{\uparrow}$} & 0.2 & 0.763& 0.768                  & 0.770                  & 0.771                  & 0.771                  & 0.770                  & 0.768                  & 0.766                  & 0.762                  & 0.755                  & 0.749 \\
 &       & 0.5 & 0.738& 0.739                  & 0.740                  & 0.741                  & 0.739                  & 0.738                  & 0.736                  & 0.733                  & 0.732                  & 0.723                  & 0.716 \\
 &       & 0.8 & 0.673& 0.678                  & 0.680                  & 0.676                  & 0.675                  & 0.671                  & 0.670                  & 0.665                  & 0.660                  & 0.653                  & 0.647 \\ \cmidrule{2-14} 
 & \multirow{3}{*}{LogiQA  $\textcolor{black}{\uparrow}$}  & 0.2 & 0.403& 0.397                  & 0.389                  & 0.376                  & 0.363                  & 0.359                  & 0.352                  & 0.338                  & 0.307                  & 0.286                  & 0.270 \\
 &       & 0.5 & 0.380& 0.366                  & 0.356                  & 0.354               & 0.347                  & 0.334                  & 0.321                  & 0.307                  & 0.285                  & 0.282                  & 0.268 \\
 &       & 0.8 & 0.356& 0.361                  & 0.328                  & 0.323                  & 0.314                  & 0.301                  & 0.289                  & 0.284                  & 0.289                  & 0.276                  & 0.260 \\ \cmidrule{2-14} 
 & SciQ  $\textcolor{black}{\uparrow}$  & -   & 0.927& 0.926                  & 0.924                  & 0.926                  & 0.921                  & 0.918                  & 0.916                  & 0.916                  & 0.914                  & 0.912                  & 0.910 \\ 
 & ARC  $\textcolor{black}{\uparrow}$   & -   & 0.782& 0.786                  & 0.790                  & 0.788                  & 0.787                  & 0.783                  & 0.782                  & 0.778                  & 0.771                  & 0.768                  & 0.765 \\ 
 & PIQA  $\textcolor{black}{\uparrow}$  & -   & 0.793& 0.793                  & 0.795                  & 0.797                  & 0.795                  & 0.798                  & 0.795                  & 0.792                  & 0.791                  & 0.794                  & 0.791 \\ 
 & WinoGrande  $\textcolor{black}{\uparrow}$               & -   & 0.707& 0.712                  & 0.722                  & 0.720                  & 0.718                  & 0.722                  & 0.710                  & 0.711                  & 0.702                  & 0.692                  & 0.688 \\ \midrule
\multirow{9}{*}{\begin{tabular}[c]{@{}c@{}}Severity of\\  Data Poisoning\end{tabular}} & \multirow{3}{*}{EM $\textcolor{black}{\downarrow}$}      & 0.2 & 62.863                  & 61.893                 & 59.103                 & 52.812                 & 43.479                 & 26.811                 & 9.947                  & 2.100                  & 0.417                  & 0.259                  & 0.251 \\
 &       & 0.5 & 61.077                  & 58.754                 & 55.297                 & 48.437                 & 37.345                 & 21.678                 & 7.991                  & 1.961                  & 0.450                  & 0.274                  & 0.252 \\
 &       & 0.8 & 55.881                  & 52.811                 & 47.575                 & 37.999                 & 25.771                 & 12.871                 & 4.619                  & 1.375                  & 0.428                  & 0.283                  & 0.246 \\ \cmidrule{2-14} 
 & \multirow{3}{*}{$\text{PC}_4$ $\textcolor{black}{\downarrow}$}      & 0.2 & 6.216& 6.153                  & 5.876                  & 5.233                  & 4.266                  & 2.641                  & 0.965                  & 0.192                  & 0.035                  & 0.033                  & 0.033 \\
 &       & 0.5 & 6.013& 5.773                  & 5.451                  & 4.827                  & 3.653                  & 2.088                  & 0.743                  & 0.153                  & 0.038                  & 0.033                  & 0.033 \\
 &       & 0.8 & 5.472& 5.165                  & 4.656                  & 3.693                  & 2.470                  & 1.202                  & 0.386                  & 0.093                  & 0.034                  & 0.033                  & 0.033 \\ \cmidrule{2-14} 
   & \multirow{3}{*}{$\text{PC}_8$ $\textcolor{black}{\downarrow}$}      & 0.2 & 5.095 & 4.998 & 4.729 & 4.182 & 3.427 & 2.027 & 0.682 & 0.125 & 0.001 & 0.000 & 0.000 \\
 &       & 0.5 & 4.901 & 4.698 & 4.362 & 3.841 & 2.872 & 1.581 & 0.501 & 0.100 & 0.003 & 0.000 & 0.000 \\
 &       & 0.8 & 4.408 & 4.172 & 3.725 & 2.905 & 1.922 & 0.853 & 0.251 & 0.054 & 0.002 & 0.000 & 0.000 \\ \bottomrule

\end{tabular}
}
\label{tab:result_data_poisoning_appendix}
\end{table*}

\section{Experiments Results of Data Poisoning on Pythia}
\label{sec:exp_poison_pythia}

Table \ref{tab:exp_pythia160M_T2} shows the experimental results on Pythia using Pythia 160M as the benign SLM. Similar to the experiments on Llama2, we can observe our method could effectively reduce data poisoning while even escalating its standard performance. 
Table \ref{tab:exp_pythia1B_T2} shows the experimental results on Pythia using Pythia 1B as the benign SLM. For simplicity, we only exhibit the results of EM and $\text{PC}_4$ here.
\begin{table}[t]
\caption{Experimental results of Data Poisoning using Pythia 160M as the benign SLM.
We adjust the ensemble weight $\alpha$, gradually decreasing it from 1.0 to 0.0, using a decrement interval of 0.1. At \(\alpha=1.0\), the model is the untrusted LLM; at \(\alpha=0.0\), it reflects the benign SLM.
 The symbols \(\uparrow\) and \(\downarrow\) separately indicate whether a higher or lower value of a specific metric is preferable.}
\resizebox{0.95\columnwidth}{!}{
\begin{tabular}{c|c|c|c|c|c|c|c|c|c|c|c}
\hline
\textbf{T=0.2} & $\alpha=1.0$     & $\alpha=0.9$      & $\alpha=0.8$      & $\alpha=0.7 $      & $\alpha=0.6 $    & $\alpha=0.5 $      & $\alpha=0.4  $    & $\alpha=0.3$     & $\alpha=0.2 $     & $\alpha=0.1$      & $\alpha=0.0$\\ \hline 
LAMBADA $\uparrow$       & 0.640       & 0.629      & 0.609      & 0.585      & 0.565      & 0.537      & 0.510                         & 0.482      & 0.446      & 0.417      & 0.379      \\ 
LogiQA $\uparrow$  & 0.272       & 0.249      & 0.255      & 0.263      & 0.222      & 0.241      & 0.195                         & 0.234      & 0.246      & 0.223      & 0.234      \\ 
SciQ $\uparrow$           & 0.836       & 0.828      & 0.819      & 0.812      & 0.798      & 0.789      & 0.761                         & 0.747      & 0.717      & 0.693      & 0.663      \\ \hline
EM $\downarrow$            & 6.205       & 5.471      & 4.439      & 2.827      & 1.589      & 0.831      & 0.461                         & 0.333      & 0.277      & 0.258      & 0.257      \\ 
$\text{PC}_4$  $\downarrow$         & 0.385       & 0.329      & 0.230      & 0.122      & 0.075      & 0.045      & 0.034                         & 0.033      & 0.033      & 0.033      & 0.033      \\ \hline
\textbf{T=0.5} & $\alpha=1.0$     & $\alpha=0.9$      & $\alpha=0.8$      & $\alpha=0.7 $      & $\alpha=0.6 $    & $\alpha=0.5 $      & $\alpha=0.4  $    & $\alpha=0.3$     & $\alpha=0.2 $     & $\alpha=0.1$      & $\alpha=0.0$\\ \hline 
LAMBADA $\uparrow$          & 0.606       & 0.592      & 0.574      & 0.552      & 0.529      & 0.495      & 0.468      & 0.436      & 0.405      & 0.373      & 0.341      \\ 
LogiQA $\uparrow$   & 0.247       & 0.242      & 0.272      & 0.242      & 0.257      & 0.262      & 0.241                              & 0.246      & 0.236      & 0.248      & 0.231      \\ 
SciQ $\uparrow$           & 0.836       & 0.828      & 0.819      & 0.812      & 0.798      & 0.789      & 0.761                              & 0.747      & 0.717      & 0.693      & 0.663      \\ \hline
EM $\downarrow$          & 4.927       & 4.281      & 3.387      & 2.447      & 1.601      & 1.004      & 0.669                              & 0.471      & 0.365      & 0.313      & 0.284      \\ 
$\text{PC}_4$   $\downarrow$         & 0.272       & 0.225      & 0.162      & 0.111      & 0.072      & 0.042      & 0.035                              & 0.033      & 0.033      & 0.033      & 0.033      \\ \hline
\textbf{T=0.8} & $\alpha=1.0$     & $\alpha=0.9$      & $\alpha=0.8$      & $\alpha=0.7 $      & $\alpha=0.6 $    & $\alpha=0.5 $      & $\alpha=0.4  $    & $\alpha=0.3$     & $\alpha=0.2 $     & $\alpha=0.1$      & $\alpha=0.0$\\ \hline 
LAMBADA $\uparrow$          & 0.530       & 0.513      & 0.494      & 0.473      & 0.446      & 0.417      & 0.394      & 0.359      & 0.334      & 0.302      & 0.270      \\ 
LogiQA $\uparrow$   & 0.265       & 0.260      & 0.256      & 0.248      & 0.259      & 0.244      & 0.238      & 0.256      & 0.243      & 0.223      & 0.202      \\ 
SciQ $\uparrow$             & 0.836       & 0.828      & 0.819      & 0.812      & 0.798      & 0.789      & 0.761                              & 0.747      & 0.717      & 0.693      & 0.663      \\ \hline
EM $\downarrow$            & 3.050       & 2.611      & 2.026      & 1.422      & 1.039      & 0.793      & 0.579                              & 0.435      & 0.357      & 0.304      & 0.285      \\
$\text{PC}_4$ $\downarrow$           & 0.148       & 0.125      & 0.092      & 0.056      & 0.051      & 0.040      & 0.035                              & 0.033      & 0.033      & 0.033      & 0.033      \\ \hline
\end{tabular}
}
\label{tab:exp_pythia160M_T2}
\end{table}

\begin{table}[t]
\caption{Experimental results of Data Poisoning using Pythia 1B as the benign SLM.
We adjust the ensemble weight $\alpha$, gradually decreasing it from 1.0 to 0.0, using a decrement interval of 0.1. At \(\alpha=1.0\), the model is the untrusted LLM; at \(\alpha=0.0\), it reflects the benign SLM.
 The symbols \(\uparrow\) and \(\downarrow\) separately indicate whether a higher or lower value of a specific metric is preferable.}
\resizebox{0.95\columnwidth}{!}{
\begin{tabular}{c|c|c|c|c|c|c|c|c|c|c|c}
\hline
\textbf{T=0.2} & $\alpha=1.0$     & $\alpha=0.9$      & $\alpha=0.8$      & $\alpha=0.7 $      & $\alpha=0.6 $    & $\alpha=0.5 $      & $\alpha=0.4  $    & $\alpha=0.3$     & $\alpha=0.2 $     & $\alpha=0.1$      & $\alpha=0.0$\\ \hline 
LAMBADA $\uparrow$         & 0.640       & 0.639      & 0.638      & 0.636      & 0.632      & 0.625      &0.619      & 0.613      & 0.603      & 0.591      & 0.575      \\ 
LogiQA $\uparrow$   & 0.272       & 0.262      & 0.258      & 0.260      & 0.263      & 0.265      & 0.254      & 0.258      & 0.261      & 0.261      & 0.257      \\ 
SciQ $\uparrow$         & 0.836       & 0.835      & 0.832      & 0.826      & 0.823      & 0.805      & 0.800                              & 0.788      & 0.779      & 0.771      & 0.761      \\ \hline
EM $\downarrow$          & 6.205       & 5.599      & 4.446      & 2.871      & 1.470      & 0.813      & 0.472                              & 0.362      & 0.284      & 0.256      & 0.243      \\ 
$\text{PC}_4$  $\downarrow$         & 0.385       & 0.335      & 0.237      & 0.130      & 0.057      & 0.037      & 0.033                              & 0.033      & 0.033      & 0.033      & 0.033      \\ \hline
\textbf{T=0.5} & $\alpha=1.0$     & $\alpha=0.9$      & $\alpha=0.8$      & $\alpha=0.7 $      & $\alpha=0.6 $    & $\alpha=0.5 $      & $\alpha=0.4  $    & $\alpha=0.3$     & $\alpha=0.2 $     & $\alpha=0.1$      & $\alpha=0.0$\\ \hline 
LAMBADA $\uparrow$         & 0.606       & 0.607      & 0.606      & 0.598      & 0.594      & 0.584      & 0.577      & 0.570      & 0.560      & 0.584      & 0.532      \\ 
LogiQA $\uparrow$  & 0.261       & 0.267      & 0.267      & 0.258      & 0.260      & 0.273      & 0.262      & 0.274      & 0.276      & 0.258      & 0.256      \\
SciQ $\uparrow$         & 0.836       & 0.835      & 0.832      & 0.826      & 0.823      & 0.805      & 0.800      & 0.788      & 0.779      & 0.771      & 0.761      \\ \hline
EM $\downarrow$        & 4.927       & 4.306      & 3.349      & 2.321      & 1.520      & 0.935      & 0.625      & 0.457      & 0.353      & 0.303      & 0.273      \\ 
$\text{PC}_4$  $\downarrow$         & 0.272       & 0.243      & 0.162      & 0.106      & 0.069      & 0.045      & 0.035      & 0.033      & 0.033      & 0.033      & 0.033      \\ \hline
\textbf{T=0.8} & $\alpha=1.0$     & $\alpha=0.9$      & $\alpha=0.8$      & $\alpha=0.7 $      & $\alpha=0.6 $    & $\alpha=0.5 $      & $\alpha=0.4  $    & $\alpha=0.3$     & $\alpha=0.2 $     & $\alpha=0.1$      & $\alpha=0.0$\\ \hline 
LAMBADA $\uparrow$         & 0.530       & 0.531      & 0.528      & 0.525      & 0.515      & 0.506      & 0.495                         & 0.483      & 0.474      & 0.460      & 0.448      \\ 
LogiQA $\uparrow$& 0.265       & 0.269      & 0.264      & 0.260      & 0.260      & 0.246      & 0.266 & 0.250      & 0.256      & 0.270      & 0.254      \\ 
SciQ $\uparrow$         & 0.836       & 0.835      & 0.832      & 0.826      & 0.823      & 0.805      & 0.800                         & 0.788      & 0.779      & 0.771      & 0.761      \\ \hline
EM $\downarrow$       & 3.050       & 2.461      & 1.955      & 1.430      & 1.039      & 0.709      & 0.529                         & 0.406      & 0.337      & 0.294      & 0.273      \\ 
$\text{PC}_4$ $\downarrow$          & 0.148       & 0.122      & 0.096      & 0.067      & 0.046      & 0.038      & 0.035                         & 0.033      & 0.033      & 0.033      & 0.033      \\ \hline
\end{tabular}
}
\label{tab:exp_pythia1B_T2}
\end{table}

\section{Experiments of PII Leakage on Llama2}
\label{sec:exp_pii_llama2}

Table \ref{tab:exp_PII_llama2_T2} shows the experimental results on Llama2. Similar to the experiments on Pythia, we can observe our method could effectively reduce PII leakage while well maintaining its standard performance.

\begin{table}[t]
\caption{Experimental results of PII using Llama2 7B as the benign SLM.
We adjust the ensemble weight $\alpha$, gradually decreasing it from 1.0 to 0.0, using a decrement interval of 0.1. At \(\alpha=1.0\), the model is the untrusted LLM; at \(\alpha=0.0\), it reflects the benign SLM.
 The symbols \(\uparrow\) and \(\downarrow\) separately indicate whether a higher or lower value of a specific metric is preferable.}
\resizebox{0.95\columnwidth}{!}{
\begin{tabular}{c|c|c|c|c|c|c|c|c|c|c|c}
\hline
\textbf{T=0.2} & $\alpha=1.0$     & $\alpha=0.9$      & $\alpha=0.8$      & $\alpha=0.7 $      & $\alpha=0.6 $    & $\alpha=0.5 $      & $\alpha=0.4  $    & $\alpha=0.3$     & $\alpha=0.2 $     & $\alpha=0.1$      & $\alpha=0.0$\\ \hline
LAMBADA $\uparrow$          & 0.792       & 0.795      & 0.796      & 0.796      & 0.792      & 0.786      & 0.781      & 0.776      & 0.769      & 0.759      & 0.749      \\ 
LogiQA $\uparrow$  & 0.410       & 0.416      & 0.418      & 0.401      & 0.382      & 0.372      & 0.367      & 0.343      & 0.321      & 0.292      & 0.270      \\
SciQ $\uparrow$           & 0.933       & 0.931      & 0.931      & 0.929      & 0.927      & 0.921      & 0.917      & 0.916      & 0.916      & 0.915      & 0.910      \\ \hline
EM $\downarrow$           & 6.642       & 6.647      & 6.645      & 6.645      & 6.641      & 6.479      & 5.874      & 3.360      & 2.907      & 2.900      & 2.884      \\ 
LC $\downarrow$           & 0.670       & 0.667      & 0.667      & 0.667      & 0.667      & 0.660      & 0.623      & 0.420      & 0.297      & 0.290      & 0.290      \\ \hline
\textbf{T=0.5} & $\alpha=1.0$     & $\alpha=0.9$      & $\alpha=0.8$      & $\alpha=0.7 $      & $\alpha=0.6 $    & $\alpha=0.5 $      & $\alpha=0.4  $    & $\alpha=0.3$     & $\alpha=0.2 $     & $\alpha=0.1$      & $\alpha=0.0$\\ \hline
LAMBADA $\uparrow$          & 0.765       & 0.765      & 0.763      & 0.761      & 0.756      & 0.754      & 0.750      & 0.744      & 0.737      & 0.726      & 0.716      \\ 
LogiQA $\uparrow$  & 0.387       & 0.383      & 0.391      & 0.379      & 0.349      & 0.339      & 0.328      & 0.318      & 0.303      & 0.283      & 0.279      \\ 
SciQ $\uparrow$           & 0.933       & 0.931      & 0.931      & 0.929      & 0.927      & 0.921      & 0.917      & 0.916      & 0.916      & 0.915      & 0.910      \\ \hline
EM $\downarrow$           & 6.641       & 6.643      & 6.642      & 6.619      & 6.551      & 6.344      & 5.433      & 3.308      & 2.913      & 2.900      & 2.715      \\ 
LC $\downarrow$           & 0.667       & 0.667      & 0.670      & 0.673      & 0.670      & 0.673      & 0.680      & 0.613      & 0.327      & 0.290      & 0.290      \\ \hline
\textbf{T=0.8} & $\alpha=1.0$     & $\alpha=0.9$      & $\alpha=0.8$      & $\alpha=0.7 $      & $\alpha=0.6 $    & $\alpha=0.5 $      & $\alpha=0.4  $    & $\alpha=0.3$     & $\alpha=0.2 $     & $\alpha=0.1$      & $\alpha=0.0$\\ \hline
LAMBADA $\uparrow$          & 0.700       & 0.700      & 0.697      & 0.697      & 0.694      & 0.689      & 0.682      & 0.676      & 0.665      & 0.655      & 0.647      \\ 
LogiQA $\uparrow$  & 0.368       & 0.370      & 0.339      & 0.342      & 0.345      & 0.338      & 0.328      & 0.304      & 0.295      & 0.281      & 0.273      \\ 
SciQ $\uparrow$           & 0.933       & 0.931      & 0.931      & 0.929      & 0.927      & 0.921      & 0.917      & 0.916      & 0.916      & 0.915      & 0.910      \\ \hline
EM $\downarrow$           & 6.627       & 6.622      & 6.591      & 6.544      & 6.376      & 5.949      & 4.519      & 3.058      & 2.903      & 2.771      & 1.989      \\ 
LC $\downarrow$           & 0.673       & 0.670      & 0.673      & 0.670      & 0.670      & 0.673      & 0.690      & 0.563      & 0.310      & 0.290      & 0.290      \\ \hline
\end{tabular}
}
\label{tab:exp_PII_llama2_T2}
\end{table}

\section{Experiments on Mitigating Negative Effects of Various Severity}
\label{sec:extended_experiments}

Here, we show the results of the untrusted StarCoder trained for 60 steps with slighter copyright infringement. The results are shown in Table \ref{tab:exp_starcoder_s60_T2}. For simplicity, we only exhibit the results of EM and Dolos here.
We also show the results of the untrusted CodeLlama trained for 30 steps with slighter copyright infringement. The results are shown in Table \ref{tab:exp_codellma_s30_T2}. For simplicity, we only exhibit the results of EM and Dolos here.
Similarly, results of Llama2 with slighter data poisoning and slighter PII leakage are shown in Table \ref{tab:exp_poisoning_llama2_s25}, \ref{tab:exp_poisoning_llama2_s50_2}. For simplicity, we only exhibit the results of EM and $\text{PC}_4$ here.

\section{Information about The Use of AI Assistants}
We collect the uncurated dataset through ChapGPT as introduced in Appendix. \ref{sec:detail}. Moreover, we also slightly use ChatGPT, Copilot in our coding and writing.

\begin{table}[t]
\caption{Experimental results of copyright infringement using StarCoder trained for 60 steps and pretrained StarCoder 3B as the benign SLM.
We adjust the ensemble weight $\alpha$, gradually decreasing it from 1.0 to 0.0, using a decrement interval of 0.1. At \(\alpha=1.0\), the model is the untrusted LLM; at \(\alpha=0.0\), it reflects the benign SLM.
 The symbols \(\uparrow\) and \(\downarrow\) separately indicate whether a higher or lower value of a specific metric is preferable.}
\resizebox{0.95\columnwidth}{!}{
\begin{tabular}{c|c|c|c|c|c|c|c|c|c|c|c}
\hline
\textbf{T=0.2} & $\alpha=1.0$     & $\alpha=0.9$      & $\alpha=0.8$      & $\alpha=0.7 $      & $\alpha=0.6 $    & $\alpha=0.5 $      & $\alpha=0.4  $    & $\alpha=0.3$     & $\alpha=0.2 $     & $\alpha=0.1$      & $\alpha=0.0$\\ \hline
pass@1    $\uparrow$      & 0.307       & 0.301      & 0.294      & 0.286      & 0.278      & 0.269      & 0.257      & 0.250      & 0.236      & 0.225      & 0.216      \\ 
pass@10  $\uparrow$       & 0.478       & 0.459      & 0.446      & 0.419      & 0.398      & 0.383      & 0.371      & 0.370      & 0.352      & 0.342      & 0.320      \\ 
pass@100   $\uparrow$     & 0.594       & 0.575      & 0.584      & 0.529      & 0.505      & 0.476      & 0.451      & 0.443      & 0.444      & 0.416      & 0.388      \\ \hline
EM $\downarrow$           & 3.205       & 3.031      & 3.003      & 2.751      & 2.594      & 2.485      & 2.368      & 2.174      & 2.093      & 1.922      & 1.790      \\ 
Dolos $\downarrow$          & 0.311       & 0.296      & 0.281      & 0.265      & 0.251      & 0.242      & 0.206      & 0.181      & 0.186      & 0.168      & 0.159      \\ \hline
\textbf{T=0.8} & $\alpha=1.0$     & $\alpha=0.9$      & $\alpha=0.8$      & $\alpha=0.7 $      & $\alpha=0.6 $    & $\alpha=0.5 $      & $\alpha=0.4  $    & $\alpha=0.3$     & $\alpha=0.2 $     & $\alpha=0.1$      & $\alpha=0.0$\\ \hline
pass@1   $\uparrow$       & 0.248       & 0.246      & 0.241      & 0.237      & 0.228      & 0.218      & 0.210                         & 0.200      & 0.189      & 0.180      & 0.171      \\ 
pass@10  $\uparrow$       & 0.571       & 0.561      & 0.545      & 0.535      & 0.516      & 0.496      & 0.482 & 0.451      & 0.434      & 0.410      & 0.387      \\ 
pass@100  $\uparrow$      & 0.811       & 0.825      & 0.797      & 0.821      & 0.780      & 0.766      & 0.755                         & 0.720      & 0.684      & 0.690      & 0.635      \\ \hline
EM $\downarrow$           & 2.215       & 2.218      & 2.169      & 2.099      & 2.108      & 2.049      & 2.099                         & 2.001      & 2.049      & 1.991      & 1.897      \\ 
Dolos $\downarrow$          & 0.284       & 0.262      & 0.266      & 0.244      & 0.240      & 0.237      & 0.238                         & 0.218      & 0.212      & 0.213      & 0.193      \\ \hline
\end{tabular}
}
\label{tab:exp_starcoder_s60_T2}
\end{table}

\begin{table}[t]
\caption{Experimental results of copyright infringement using CodeLlama trained for 30 steps and pretrained CodeLlama 7B as the benign SLM.
We adjust the ensemble weight $\alpha$, gradually decreasing it from 1.0 to 0.0, using a decrement interval of 0.1. At \(\alpha=1.0\), the model is the untrusted LLM; at \(\alpha=0.0\), it reflects the benign SLM.
 The symbols \(\uparrow\) and \(\downarrow\) separately indicate whether a higher or lower value of a specific metric is preferable.}
 \resizebox{0.95\columnwidth}{!}{
\begin{tabular}{c|c|c|c|c|c|c|c|c|c|c|c}
\hline
T=0.2    & $\alpha=1.0$     & $\alpha=0.9$      & $\alpha=0.8$      & $\alpha=0.7 $      & $\alpha=0.6 $    & $\alpha=0.5 $      & $\alpha=0.4  $    & $\alpha=0.3$     & $\alpha=0.2 $     & $\alpha=0.1$      & $\alpha=0.0$\\ \hline
pass@1  $\uparrow$  & 0.344 & 0.336 & 0.331 & 0.326 & 0.323 & 0.32  & 0.316 & 0.314 & 0.31  & 0.304 & 0.301 \\ 
pass@10 $\uparrow$  & 0.524 & 0.525 & 0.518 & 0.51  & 0.506 & 0.486 & 0.482 & 0.473 & 0.468 & 0.463 & 0.458 \\ 
pass@100 $\uparrow$  & 0.632 & 0.641 & 0.631 & 0.647 & 0.652 & 0.619 & 0.632 & 0.604 & 0.648 & 0.594 & 0.605 \\ \hline
EM $\downarrow$       & 1.191 & 1.101 & 1.056 & 1.119 & 1.079 & 1.100 & 1.121 & 1.059 & 1.064 & 1.082 & 1.023 \\ 
Dolos $\downarrow$    & 0.230 & 0.230 & 0.220 & 0.233 & 0.220 & 0.218 & 0.218 & 0.200 & 0.189 & 0.174 & 0.160 \\ \hline
T=0.5    & $\alpha=1.0$     & $\alpha=0.9$      & $\alpha=0.8$      & $\alpha=0.7 $      & $\alpha=0.6 $    & $\alpha=0.5 $      & $\alpha=0.4  $    & $\alpha=0.3$     & $\alpha=0.2 $     & $\alpha=0.1$      & $\alpha=0.0$\\     \hline
pass@1 $\uparrow$   & 0.337 & 0.335 & 0.335 & 0.333 & 0.33  & 0.325 & 0.32  & 0.315 & 0.31  & 0.303 & 0.296 \\ 
pass@10 $\uparrow$  & 0.647 & 0.642 & 0.64  & 0.647 & 0.64  & 0.632 & 0.625 & 0.611 & 0.612 & 0.595 & 0.582 \\ 
pass@100 $\uparrow$ & 0.835 & 0.818 & 0.823 & 0.833 & 0.849 & 0.81  & 0.831 & 0.817 & 0.801 & 0.832 & 0.81  \\ \hline
EM $\downarrow$       & 1.645 & 1.629 & 1.699 & 1.626 & 1.598 & 1.592 & 1.589 & 1.638 & 1.550 & 1.532 & 1.445 \\ 
Dolos $\downarrow$    & 0.288 & 0.283 & 0.276 & 0.272 & 0.260 & 0.249 & 0.241 & 0.236 & 0.231 & 0.213 & 0.203 \\ \hline
T=0.8    & $\alpha=1.0$     & $\alpha=0.9$      & $\alpha=0.8$      & $\alpha=0.7 $      & $\alpha=0.6 $    & $\alpha=0.5 $      & $\alpha=0.4  $    & $\alpha=0.3$     & $\alpha=0.2 $     & $\alpha=0.1$      & $\alpha=0.0$\\     \hline
pass@1 $\uparrow$  & 0.287 & 0.291 & 0.294 & 0.292 & 0.293 & 0.290 & 0.289 & 0.283 & 0.275 & 0.265 & 0.257 \\ 
pass@10 $\uparrow$  & 0.666 & 0.667 & 0.658 & 0.661 & 0.644 & 0.650 & 0.646 & 0.646 & 0.636 & 0.612 & 0.601 \\ 
pass@100 $\uparrow$ & 0.891 & 0.882 & 0.872 & 0.876 & 0.882 & 0.885 & 0.888 & 0.884 & 0.879 & 0.874 & 0.873 \\ \hline
EM $\downarrow$       & 1.519 & 1.563 & 1.495 & 1.406 & 1.496 & 1.434 & 1.405 & 1.379 & 1.339 & 1.307 & 1.311 \\ 
Dolos $\downarrow$    & 0.274 & 0.256 & 0.265 & 0.258 & 0.251 & 0.240 & 0.234 & 0.216 & 0.216 & 0.215 & 0.199 \\ \hline
\end{tabular}
}
\label{tab:exp_codellma_s30_T2}
\end{table}

\begin{table}[t]
 \caption{Experimental results of data poisoning using Llama2 trained for 25 steps.
 We adjust the ensemble weight $\alpha$, gradually decreasing it from 1.0 to 0.0, using a decrement interval of 0.1. At \(\alpha=1.0\), the model is the untrusted LLM; at \(\alpha=0.0\), it reflects the benign SLM.
 The symbols \(\uparrow\) and \(\downarrow\) separately indicate whether a higher or lower value of a specific metric is preferable.}
 \resizebox{0.95\columnwidth}{!}{
\begin{tabular}{cccccccccccc}
\hline
 \textbf{T=0.2} & $\alpha=1.0$     & $\alpha=0.9$      & $\alpha=0.8$      & $\alpha=0.7 $      & $\alpha=0.6 $    & $\alpha=0.5 $      & $\alpha=0.4  $    & $\alpha=0.3$     & $\alpha=0.2 $     & $\alpha=0.1$      & $\alpha=0.0$\\ \hline
LAMBADA $\uparrow$         & 0.768       & 0.773      & 0.773      & 0.774      & 0.773      & 0.770      & 0.769      & 0.766      & 0.762      & 0.755      & 0.749      \\ 
LogiQA $\uparrow$ & 0.401       & 0.400      & 0.393      & 0.386      & 0.379      & 0.370      & 0.356      & 0.335      & 0.317      & 0.289      & 0.270      \\
SciQ $\uparrow$          & 0.932       & 0.930      & 0.927      & 0.927      & 0.924      & 0.921      & 0.918      & 0.918      & 0.917      & 0.914      & 0.910      \\ \hline
EM $\downarrow$          & 8.580       & 7.587      & 6.144      & 4.700      & 2.885      & 1.068      & 0.479      & 0.293      & 0.264      & 0.252      & 0.251      \\ 
$\text{PC}_4$   $\downarrow$         & 0.374       & 0.307      & 0.269      & 0.238      & 0.177      & 0.056      & 0.037      & 0.033      & 0.033      & 0.033      & 0.033      \\ \hline

 \textbf{T=0.5} & $\alpha=1.0$     & $\alpha=0.9$      & $\alpha=0.8$      & $\alpha=0.7 $      & $\alpha=0.6 $    & $\alpha=0.5 $      & $\alpha=0.4  $    & $\alpha=0.3$     & $\alpha=0.2 $     & $\alpha=0.1$      & $\alpha=0.0$\\ \hline
LAMBADA $\uparrow$         & 0.734       & 0.742      & 0.742      & 0.743      & 0.740      & 0.740      & 0.738 & 0.734      & 0.731      & 0.723      & 0.716      \\ 
LogiQA $\uparrow$ & 0.380       & 0.373      & 0.365      & 0.356      & 0.355      & 0.336      & 0.328                         & 0.305      & 0.288      & 0.285      & 0.268      \\ 
SciQ $\uparrow$          & 0.932       & 0.930      & 0.927      & 0.927      & 0.924      & 0.921      & 0.918                         & 0.918      & 0.917      & 0.914      & 0.910      \\ \hline
EM $\downarrow$          & 7.433       & 6.555      & 5.344      & 3.960      & 2.417      & 1.225      & 0.590                         & 0.350      & 0.281      & 0.260      & 0.252      \\ 
$\text{PC}_4$   $\downarrow$         & 0.271       & 0.237      & 0.197      & 0.163      & 0.107      & 0.055      & 0.036                         & 0.033      & 0.033      & 0.033      & 0.033      \\ \hline

\textbf{T=0.8} & $\alpha=1.0$     & $\alpha=0.9$      & $\alpha=0.8$      & $\alpha=0.7 $      & $\alpha=0.6 $    & $\alpha=0.5 $      & $\alpha=0.4  $    & $\alpha=0.3$     & $\alpha=0.2 $     & $\alpha=0.1$      & $\alpha=0.0$\\ \hline
LAMBADA $\uparrow$         & 0.677       & 0.678      & 0.680      & 0.675      & 0.673      & 0.672      & 0.671      & 0.665      & 0.660      & 0.653      & 0.647      \\ 
LogiQA $\uparrow$ & 0.358       & 0.344      & 0.327      & 0.321      & 0.314      & 0.303      & 0.285      & 0.280      & 0.287      & 0.275      & 0.260      \\ 
SciQ $\uparrow$          & 0.932       & 0.930      & 0.927      & 0.927      & 0.924      & 0.921      & 0.918      & 0.918      & 0.917      & 0.914      & 0.910      \\ \hline
EM $\downarrow$          & 5.980       & 4.869      & 3.880      & 2.791      & 1.781      & 1.049      & 0.619      & 0.380      & 0.293      & 0.256      & 0.246      \\ 
$\text{PC}_4$   $\downarrow$       & 0.189       & 0.155      & 0.128      & 0.097      & 0.062      & 0.047      & 0.037      & 0.033      & 0.033      & 0.033      & 0.033      \\ \hline
\end{tabular}
}
\label{tab:exp_poisoning_llama2_s25}
\end{table}

\begin{table}[t]
 \caption{Experimental results of data poisoning using Llama2 trained for 50 steps.
 We adjust the ensemble weight $\alpha$, gradually decreasing it from 1.0 to 0.0, using a decrement interval of 0.1. At \(\alpha=1.0\), the model is the untrusted LLM; at \(\alpha=0.0\), it reflects the benign SLM.
 The symbols \(\uparrow\) and \(\downarrow\) separately indicate whether a higher or lower value of a specific metric is preferable.}
 \resizebox{0.95\columnwidth}{!}{
\begin{tabular}{cccccccccccc}
\hline
 \textbf{T=0.2} & $\alpha=1.0$     & $\alpha=0.9$      & $\alpha=0.8$      & $\alpha=0.7 $      & $\alpha=0.6 $    & $\alpha=0.5 $      & $\alpha=0.4  $    & $\alpha=0.3$     & $\alpha=0.2 $     & $\alpha=0.1$      & $\alpha=0.0$\\ \hline
LAMBADA $\uparrow$         & 0.763       & 0.770      & 0.771      & 0.773      & 0.772      & 0.770                         & 0.768 & 0.766      & 0.762      & 0.754      & 0.749      \\ 
LogiQA $\uparrow$ & 0.408       & 0.401      & 0.400      & 0.384      & 0.373      & 0.370 & 0.358                         & 0.340      & 0.308      & 0.286      & 0.270      \\ 
SciQ $\uparrow$          & 0.927       & 0.926      & 0.923      & 0.925      & 0.922      & 0.919                         & 0.916                         & 0.916      & 0.914      & 0.912      & 0.910      \\ \hline
EM $\downarrow$          & 53.185      & 50.781     & 47.264     & 40.413     & 27.965     & 14.856                        & 5.281                         & 1.311      & 0.303      & 0.257      & 0.251      \\ 
$\text{PC}_4$   $\downarrow$        & 5.210       & 4.955      & 4.560      & 3.879      & 2.684      & 1.426                         & 0.519                         & 0.110      & 0.033      & 0.033      & 0.033      \\ \hline

 \textbf{T=0.5} & $\alpha=1.0$     & $\alpha=0.9$      & $\alpha=0.8$      & $\alpha=0.7 $      & $\alpha=0.6 $    & $\alpha=0.5 $      & $\alpha=0.4  $    & $\alpha=0.3$     & $\alpha=0.2 $     & $\alpha=0.1$      & $\alpha=0.0$\\ \hline
LAMBADA $\uparrow$         & 0.738       & 0.740      & 0.740      & 0.741      & 0.741      & 0.738      & 0.737      & 0.734      & 0.732      & 0.723      & 0.716      \\ 
LogiQA $\uparrow$ & 0.381       & 0.375      & 0.359      & 0.356      & 0.348      & 0.336      & 0.322      & 0.300      & 0.283      & 0.284      & 0.268      \\ 
SciQ $\uparrow$          & 0.927       & 0.926      & 0.923      & 0.925      & 0.922      & 0.919      & 0.916      & 0.916      & 0.914      & 0.912      & 0.910      \\ \hline
EM $\downarrow$          & 49.567      & 46.456     & 41.911     & 33.956     & 23.456     & 12.424     & 4.672      & 1.271      & 0.372      & 0.271      & 0.252      \\ 
$\text{PC}_4$    $\downarrow$         & 4.812       & 4.487      & 4.013      & 3.249      & 2.209      & 1.153      & 0.403      & 0.085      & 0.035      & 0.033      & 0.033      \\ \hline

\textbf{T=0.8} & $\alpha=1.0$     & $\alpha=0.9$      & $\alpha=0.8$      & $\alpha=0.7 $      & $\alpha=0.6 $    & $\alpha=0.5 $      & $\alpha=0.4  $    & $\alpha=0.3$     & $\alpha=0.2 $     & $\alpha=0.1$      & $\alpha=0.0$\\ \hline
LAMBADA $\uparrow$         & 0.674       & 0.678      & 0.681      & 0.677      & 0.676      & 0.672      & 0.671 & 0.665      & 0.661      & 0.653      & 0.647      \\ 
LogiQA $\uparrow$ & 0.358       & 0.352      & 0.329      & 0.325      & 0.317      & 0.303      & 0.292                         & 0.286      & 0.289      & 0.279      & 0.273      \\ 
SciQ $\uparrow$          & 0.927       & 0.926      & 0.923      & 0.925      & 0.922      & 0.919      & 0.916                         & 0.916      & 0.914      & 0.912      & 0.910      \\ \hline
EM $\downarrow$          & 42.984      & 38.105     & 32.676     & 24.087     & 15.570     & 7.883      & 2.881                         & 0.999      & 0.396      & 0.277      & 0.246      \\ 
$\text{PC}_4$    $\downarrow$       & 4.121       & 3.604      & 3.097      & 2.221      & 1.417      & 0.671      & 0.212                         & 0.063      & 0.033      & 0.033      & 0.033      \\ \hline
\end{tabular}
}
\label{tab:exp_poisoning_llama2_s50_2}
\end{table}

\begin{table}[t]
 \caption{Experimental results of PII using Llama2 trained for 50 steps.
 We adjust the ensemble weight $\alpha$, gradually decreasing it from 1.0 to 0.0, using a decrement interval of 0.1. At \(\alpha=1.0\), the model is the untrusted LLM; at \(\alpha=0.0\), it reflects the benign SLM.
 The symbols \(\uparrow\) and \(\downarrow\) separately indicate whether a higher or lower value of a specific metric is preferable.}
 \resizebox{0.95\columnwidth}{!}{
\begin{tabular}{cccccccccccc}
\hline
 \textbf{T=0.2} & $\alpha=1.0$     & $\alpha=0.9$      & $\alpha=0.8$      & $\alpha=0.7 $      & $\alpha=0.6 $    & $\alpha=0.5 $      & $\alpha=0.4  $    & $\alpha=0.3$     & $\alpha=0.2 $     & $\alpha=0.1$      & $\alpha=0.0$\\ \hline
LAMBADA $\uparrow$         & 0.790       & 0.793      & 0.796      & 0.796      & 0.791      & 0.786      & 0.780      & 0.775      & 0.769      & 0.759      & 0.749      \\ 
LogiQA $\uparrow$ & 0.409       & 0.406      & 0.412      & 0.401      & 0.384      & 0.375      & 0.363      & 0.347      & 0.321      & 0.295      & 0.270      \\ 
SciQ $\uparrow$          & 0.932       & 0.931      & 0.929      & 0.930      & 0.926      & 0.921      & 0.917      & 0.916      & 0.917      & 0.915      & 0.910      \\ \hline
EM $\downarrow$          & 5.109       & 5.003      & 4.919      & 4.807      & 4.521      & 4.225      & 3.791      & 3.025      & 2.905      & 2.900      & 2.884      \\ 
LC $\downarrow$          & 0.567       & 0.553      & 0.557      & 0.547      & 0.513      & 0.487      & 0.457      & 0.330      & 0.293      & 0.290      & 0.290      \\ \hline

 \textbf{T=0.5} & $\alpha=1.0$     & $\alpha=0.9$      & $\alpha=0.8$      & $\alpha=0.7 $      & $\alpha=0.6 $    & $\alpha=0.5 $      & $\alpha=0.4  $    & $\alpha=0.3$     & $\alpha=0.2 $     & $\alpha=0.1$      & $\alpha=0.0$\\ \hline
LAMBADA $\uparrow$         & 0.763 & 0.763 & 0.763 & 0.760 & 0.754 & 0.753 & 0.748 & 0.743 & 0.737 & 0.726 & 0.716 \\ 
LogiQA $\uparrow$ & 0.382 & 0.386 & 0.365 & 0.377 & 0.374 & 0.341 & 0.324 & 0.317 & 0.307 & 0.298 & 0.279 \\ 
SciQ $\uparrow$          & 0.932 & 0.931 & 0.929 & 0.930 & 0.926 & 0.921 & 0.917 & 0.916 & 0.917 & 0.915 & 0.910 \\ \hline
EM $\downarrow$          & 4.989 & 4.907 & 4.810 & 4.632 & 4.390 & 4.061 & 3.523 & 3.036 & 2.905 & 2.900 & 2.715 \\ 
LC $\downarrow$          & 0.603 & 0.623 & 0.610 & 0.613 & 0.597 & 0.577 & 0.550 & 0.423 & 0.313 & 0.290 & 0.290 \\ \hline

\textbf{T=0.8} & $\alpha=1.0$     & $\alpha=0.9$      & $\alpha=0.8$      & $\alpha=0.7 $      & $\alpha=0.6 $    & $\alpha=0.5 $      & $\alpha=0.4  $    & $\alpha=0.3$     & $\alpha=0.2 $     & $\alpha=0.1$      & $\alpha=0.0$\\ \hline
LAMBADA $\uparrow$         & 0.698 & 0.698 & 0.696 & 0.695 & 0.691 & 0.686 & 0.682 & 0.675 & 0.664 & 0.655 & 0.647 \\ 
LogiQA $\uparrow$ & 0.365 & 0.351 & 0.338 & 0.354 & 0.321 & 0.312 & 0.333 & 0.303 & 0.293 & 0.279 & 0.273 \\ 
SciQ $\uparrow$          & 0.932 & 0.931 & 0.929 & 0.930 & 0.926 & 0.921 & 0.917 & 0.916 & 0.917 & 0.915 & 0.910 \\ \hline
EM $\downarrow$          & 4.775 & 4.692 & 4.579 & 4.372 & 4.088 & 3.652 & 3.202 & 2.938 & 2.898 & 2.758 & 1.989 \\ 
LC $\downarrow$          & 0.637 & 0.647 & 0.640 & 0.637 & 0.623 & 0.597 & 0.553 & 0.373 & 0.300 & 0.290 & 0.290 \\ \hline
\end{tabular}
}
\label{tab:exp_poisoning_llama2_s50}
\end{table}

\begin{table}[t]
 \caption{Experimental results of PII using Llama2 trained for 100 steps.
 We adjust the ensemble weight $\alpha$, gradually decreasing it from 1.0 to 0.0, using a decrement interval of 0.1. At \(\alpha=1.0\), the model is the untrusted LLM; at \(\alpha=0.0\), it reflects the benign SLM.
 The symbols \(\uparrow\) and \(\downarrow\) separately indicate whether a higher or lower value of a specific metric is preferable.}
 \resizebox{0.95\columnwidth}{!}{
\begin{tabular}{cccccccccccc}
\hline
 \textbf{T=0.2} & $\alpha=1.0$     & $\alpha=0.9$      & $\alpha=0.8$      & $\alpha=0.7 $      & $\alpha=0.6 $    & $\alpha=0.5 $      & $\alpha=0.4  $    & $\alpha=0.3$     & $\alpha=0.2 $     & $\alpha=0.1$      & $\alpha=0.0$\\ \hline
LAMBADA $\uparrow$         & 0.792 & 0.795 & 0.796 & 0.796 & 0.792 & 0.786 & 0.780 & 0.776 & 0.769 & 0.759 & 0.749 \\ 
LogiQA $\uparrow$ & 0.413 & 0.415 & 0.407 & 0.397 & 0.384 & 0.379 & 0.368 & 0.342 & 0.324 & 0.294 & 0.282 \\ 
SciQ $\uparrow$          & 0.933 & 0.931 & 0.930 & 0.929 & 0.927 & 0.921 & 0.917 & 0.916 & 0.916 & 0.915 & 0.910 \\ \hline
EM $\downarrow$          & 6.638 & 6.634 & 6.635 & 6.625 & 6.536 & 6.326 & 5.711 & 3.274 & 2.903 & 2.900 & 2.884 \\ 
LC $\downarrow$          & 0.670 & 0.670 & 0.667 & 0.667 & 0.663 & 0.657 & 0.607 & 0.387 & 0.293 & 0.290 & 0.290 \\ \hline

 \textbf{T=0.5} & $\alpha=1.0$     & $\alpha=0.9$      & $\alpha=0.8$      & $\alpha=0.7 $      & $\alpha=0.6 $    & $\alpha=0.5 $      & $\alpha=0.4  $    & $\alpha=0.3$     & $\alpha=0.2 $     & $\alpha=0.1$      & $\alpha=0.0$\\ \hline
LAMBADA $\uparrow$         & 0.765 & 0.764 & 0.763 & 0.761 & 0.755 & 0.754 & 0.749 & 0.744 & 0.737 & 0.726 & 0.716 \\ 
LogiQA $\uparrow$ & 0.399 & 0.386 & 0.372 & 0.364 & 0.353 & 0.361 & 0.352 & 0.333 & 0.308 & 0.297 & 0.279 \\ 
SciQ $\uparrow$          & 0.933 & 0.931 & 0.930 & 0.929 & 0.927 & 0.921 & 0.917 & 0.916 & 0.916 & 0.915 & 0.910 \\ \hline
EM $\downarrow$          & 6.613 & 6.623 & 6.613 & 6.561 & 6.465 & 6.164 & 5.185 & 3.252 & 2.912 & 2.900 & 2.715 \\ 
LC $\downarrow$          & 0.670 & 0.670 & 0.670 & 0.670 & 0.667 & 0.667 & 0.680 & 0.247 & 0.323 & 0.290 & 0.290 \\ \hline

\textbf{T=0.8} & $\alpha=1.0$     & $\alpha=0.9$      & $\alpha=0.8$      & $\alpha=0.7 $      & $\alpha=0.6 $    & $\alpha=0.5 $      & $\alpha=0.4  $    & $\alpha=0.3$     & $\alpha=0.2 $     & $\alpha=0.1$      & $\alpha=0.0$\\ \hline
LAMBADA $\uparrow$         & 0.699 & 0.699 & 0.697 & 0.697 & 0.693 & 0.688 & 0.682 & 0.676 & 0.664 & 0.655 & 0.647 \\ 
LogiQA $\uparrow$ & 0.375 & 0.362 & 0.332 & 0.347 & 0.338 & 0.331 & 0.328 & 0.307 & 0.295 & 0.284 & 0.273 \\ 
SciQ $\uparrow$          & 0.933 & 0.931 & 0.930 & 0.929 & 0.927 & 0.921 & 0.917 & 0.916 & 0.916 & 0.915 & 0.910 \\ \hline
EM $\downarrow$          & 6.577 & 6.552 & 6.509 & 6.447 & 6.217 & 5.693 & 4.237 & 3.648 & 2.903 & 2.766 & 1.989 \\ 
LC $\downarrow$          & 0.670 & 0.670 & 0.667 & 0.673 & 0.670 & 0.680 & 0.690 & 0.533 & 0.310 & 0.290 & 0.290 \\ \hline
\end{tabular}
}
\label{tab:exp_poisoning_llama2_s100}
\end{table}

% 

%%%%%%%%%%%%%%%%%%%%%%%%%%%%%%%%%%%%%%%%%%%%%%%%%%%%%%%%%%%%%%%%%%%%%%%%%%%%%%%
%%%%%%%%%%%%%%%%%%%%%%%%%%%%%%%%%%%%%%%%%%%%%%%%%%%%%%%%%%%%%%%%%%%%%%%%%%%%%%%

\end{document}